\documentclass{article}

\usepackage{microtype}
\usepackage{graphicx}
\usepackage{subcaption}
\usepackage{booktabs} % for professional tables

\usepackage{hyperref}

\usepackage[preprint]{icml2026}

\usepackage{amsmath}
\usepackage{amssymb}
\usepackage{mathtools}
\usepackage{amsthm}

\usepackage[capitalize,noabbrev]{cleveref}

\newcommand{\mb}[1]{\mathbf{#1}}

\usepackage{tikz}
\usetikzlibrary{positioning, arrows.meta, calc, shapes.geometric, fit, backgrounds, decorations.pathreplacing}

\pgfdeclarelayer{background}
\pgfsetlayers{background,main}

\theoremstyle{plain}

\theoremstyle{definition}

\theoremstyle{remark}

\usepackage[textsize=tiny]{todonotes}

\icmltitlerunning{Superposition unifies power-law training dynamics}

\begin{document}

\twocolumn[
  \icmltitle{Superposition unifies power-law training dynamics}



  \icmlsetsymbol{equal}{*}

  \begin{icmlauthorlist}
    \icmlauthor{Zixin Jessie Chen}{mit_phys}
    \icmlauthor{Hao Chen}{pton_eecs}
    \icmlauthor{Yizhou Liu}{mit_me}
    \icmlauthor{Jeff Gore}{mit_phys}
  \end{icmlauthorlist}

  \icmlaffiliation{mit_phys}{Department of Physics, Massachusetts Institute of Technology, Cambridge, Massachusetts 02139, USA}
  \icmlaffiliation{pton_eecs}{Department of Electrical and Computer Engineering, Princeton University, Princeton, New Jersey 08544, USA}
  \icmlaffiliation{mit_me}{Department of Mechanical Engineering, Massachusetts Institute of Technology, Cambridge, Massachusetts 02139, USA}

  \icmlcorrespondingauthor{Zixin Jessie Chen}{jzxchen@mit.edu}

  \icmlkeywords{Superposition, Training Dynamics, Neural Scaling Laws, Large Language Models, Teacher-Student Model}

  \vskip 0.3in
]



\printAffiliationsAndNotice{}  % no special notice (required even if empty)

\begin{abstract}
    %%% ABSTRACT %%%

We investigate the role of feature superposition in the emergence of power-law training dynamics using a teacher-student framework. We first derive an analytic theory for training without superposition, establishing that the power-law training exponent depends on both the input data statistics and channel importance. Remarkably, we discover that a superposition bottleneck induces a transition to a universal power-law exponent of $\sim 1$, independent of data and channel statistics. This one over time training with superposition represents an up to tenfold acceleration compared to the purely sequential learning that takes place in the absence of superposition. Our finding that superposition leads to rapid training with a data-independent power law exponent may have important implications for a wide range of neural networks that employ superposition, including production-scale large language models.

\end{abstract}

%%%% INTRODUCTION %%%
\section{Introduction} \label{sec:intro}

The remarkable success of large language models (LLMs) is underpinned by \textit{neural scaling laws}, where model performance scales predictably as a power law with compute, dataset size, and parameter count \citep{kaplan2020scaling, hoffmann2022training, zhai2022scaling, clark2022unified}. These laws describe not only the final performance but also the trajectory of optimization itself, where the training loss $\mathcal{L}(t)$ often decays as $\mathcal{L}(t) \propto t^{-\alpha}$ over orders of magnitude in training steps. Despite the ubiquity of these dynamics across modalities \citep{henighan2020scaling} and architectures \citep{bi2024mamba}, the microscopic mechanisms that govern this macroscopic behavior remain a subject of intense debate.

\begin{figure}[h]
\centering
\resizebox{\linewidth}{!}{% Resize to fill width
\begin{tikzpicture}[scale=1.1, >={Latex[width=3mm,length=3mm]}, font=\small]
    % --- Panel A: Orthogonal ---
    \begin{scope}[shift={(0,0)}]
        % Grid/Axes (Darker Lines)
        \draw[step=1cm,gray!40,thin] (-2.5,-2.5) grid (2.5,2.5);
        \draw[->,thick,gray] (-2.5,0) -- (2.5,0);
        \draw[->,thick,gray] (0,-2.5) -- (0,2.5);
        
        % Features
        \draw[->, ultra thick, color=blue!70!black] (0,0) -- (2,0) 
            node[anchor=north west, xshift=2pt, font=\bfseries\large] {Feature 1};
        \draw[->, ultra thick, color=red!70!black] (0,0) -- (0,2) 
            node[anchor=south west, yshift=2pt, font=\bfseries\large] {Feature 2};
        
        % Annotation
        \node[align=center, fill=white, inner sep=3pt, draw=gray!30, rounded corners] at (0, -3.2) {Features are orthogonal to each other.\\Zero Interference};

        % NEW LABEL POSITION (Below)
        \node[font=\bfseries\Large] at (0, -4.2) {(a) No superposition};
    \end{scope}

    % --- Panel B: Superposition ---
    \begin{scope}[shift={(7.5,0)}]
        % Grid/Axes (Darker Lines)
        \draw[step=1cm,gray!40,thin] (-2.5,-2.5) grid (2.5,2.5);
        \draw[dashed, gray!60] (0,0) circle (2cm); 
        
        % Features (Pentagon)
        \foreach \angle/\col/\lab/\pos in {
            90/red!70!black/Feature 1/above, 
            162/green!60!black/Feature 2/left, 
            234/purple!70!black/Feature 3/left, 
            306/orange!80!black/Feature 4/right, 
            18/blue!70!black/Feature 5/right} 
        {
            \draw[->, ultra thick, color=\col] (0,0) -- (\angle:2cm);
            \node[color=\col, font=\bfseries\large, yshift=5pt] at (\angle:2.4cm) {\lab};
        }

        % Interference Visualization
        \draw[dashed, very thick, darkgray] (18:2cm) -- (0, {2*sin(18)}); 
        
        % Brace and Label
        \draw[very thick, red, decoration={brace, mirror, amplitude=6pt, raise=2pt}, decorate] (0,0) -- (0, {2*sin(18)}) 
            node[pos=0.5, left=-50pt, yshift=20pt, align=right, font=\bfseries, color=black!70!black] {Interference};
            
        % Annotation
        \node[align=center, fill=white, inner sep=3pt, draw=gray!30, rounded corners] at (0, -3.2) {$N > K$ Features packed. \\Non-zero Interference};

        % NEW LABEL POSITION (Below)
        \node[font=\bfseries\Large] at (0, -4.2) {(b) Superposition};
    \end{scope}
\end{tikzpicture}
}
\caption{\textbf{Geometric illustration of superposition.} \textbf{(a)} In the absence of superposition, features have zero interference with each other. \textbf{(b)} Superposition compresses features into a smaller latent size, introducing interference among them (e.g., Feature 5 projects onto Feature 1).}
\label{fig:superposition_concept}
\end{figure}
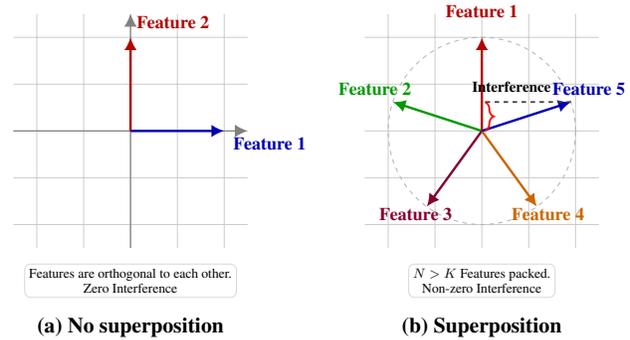

\begin{figure*}[t]
\centering
\resizebox{\textwidth}{!}{% Scale to full text width
\begin{tikzpicture}[
    % Tighter layout: reduced vertical distance
    node distance=1.5cm and 3.0cm,
    >={Latex[width=5mm,length=5mm]}, % Bigger arrows
    % Bigger blocks and fonts
    block/.style={draw, rectangle, rounded corners, minimum height=4.0cm, minimum width=3.2cm, align=center, fill=gray!5, ultra thick, font=\Large},
    studentblock/.style={draw, rectangle, rounded corners, minimum height=2.8cm, minimum width=2.8cm, align=center, fill=blue!5, ultra thick, font=\Large},
    op/.style={draw, circle, minimum size=1.5cm, fill=white, ultra thick, font=\huge},
    connector/.style={draw, circle, minimum size=1.8cm, fill=gray!10, ultra thick, font=\huge\bfseries},
    % Labels
    labeltext/.style={font=\large\sffamily, color=gray!80!black, align=center, fill=white, inner sep=4pt}
]

% --- Nodes ---

% Input
\node[connector] (input) {$\mathbf{x}$};
\node[below=0.4cm of input, labeltext] {\Large \textbf{Input}\\ \Large Dim $N$};

% Split point
\coordinate[right=2.5cm of input] (split);

% --- Student Branch (Bottom) ---
% DEFINED FIRST to establish the maximum length
% Increased vertical gap (below right=1.8cm)
\node[studentblock, below right=1.8cm and 1.5cm of split, name=encoder] {$\mathbf{W}$\\ \textbf{Embed}\\ $K \times N$};

\node[studentblock, right=2.0cm of encoder, name=student, fill=blue!15] {$\mathbf{B}$\\ \textbf{Student}\\ $K \times K$};

\node[studentblock, right=2.0cm of student, name=decoder] {$\mathbf{W}^\top$\\ \textbf{Unembed}\\ $N \times K$};

% Bias Addition
\node[op, right=1.5cm of decoder] (bias) {$+$};
\node[above=1.0cm of bias, font=\huge] (bias_vec) {$\mathbf{b}$};
\draw[->, ultra thick] (bias_vec) -- (bias);

% ReLU
\node[studentblock, right=1.5cm of bias, minimum width=2.2cm, minimum height=1.8cm, font=\bfseries\Large] (relu) {ReLU};

% Student Output y_hat
\node[connector, right=1.5cm of relu] (y_hat) {$\mathbf{y}$};

% --- Teacher Branch (Top) ---
% Increased vertical gap (above right=0.8cm)
\node[block, above right=0.8cm and 4.5cm of split, name=teacher, fill=red!5] {
    $\mathbf{A}$\\ \textbf{Teacher}\\ $N \times N$ \\
    \textit{(Channel}\\
    \textit{Importance)}\\
};

% ALIGNMENT FIX: Position y_star at the same X-coordinate as y_hat
\node[connector] (y_star) at (teacher -| y_hat) {$\mathbf{y}^*$};

% --- The Background Layer (Bottleneck Highlight) ---
\begin{pgfonlayer}{background}
    \node[draw=blue!40, dashed, fill=blue!2, fill opacity=0.4, 
          fit=(encoder) (student) (decoder), 
          inner sep=0.8cm, rounded corners=20pt, line width=2pt] (bottleneck) {};
    \node[blue!60!black, font=\bfseries\LARGE, below=0.3cm of bottleneck] {Superposition Bottleneck ($K \leq N$)};
\end{pgfonlayer}

% --- Loss ---
% Positioned to the RIGHT of the aligned outputs
% Reduced size slightly (2.2cm) to save space
\node[draw, circle, minimum size=2.2cm, right=3.0cm of $(y_star)!0.5!(y_hat)$, fill=red!10, ultra thick, font=\bfseries\huge] (loss) {MSE};

% --- Arrows ---
\draw[->, line width=2.5pt] (input) -- (split);

% Teacher Path
\draw[->, line width=2.5pt] (split) |- (teacher.west) node[pos=0.75, above, labeltext] {\Large Dim $N$};
\draw[->, line width=2.5pt] (teacher) -- (y_star) node[midway, above, font=\LARGE] {$\mathbf{Ax}$, Dim $N$};

% Student Path
\draw[->, line width=2.5pt] (split) |- (encoder.west) node[pos=0.60, below, labeltext] {\Large Dim \\ \Large $N$};
\draw[->, line width=2.5pt] (encoder) -- (student) node[midway, above, labeltext] {\Large Latent \\ $\mathbf{K}$};
\draw[->, line width=2.5pt] (student) -- (decoder) node[midway, above, labeltext] {\Large Latent \\ $\mathbf{K}$};
\draw[->, line width=2.5pt] (decoder) -- (bias);
\draw[->, line width=2.5pt] (bias) -- (relu);
\draw[->, line width=2.5pt] (relu) -- (y_hat) node[midway, above, font=\LARGE] {$\hat{\mathbf{y}}$};

% Loss connections
\draw[->, line width=2.5pt, dashed] (y_star) -| (loss);
\draw[->, line width=2.5pt, dashed] (y_hat) -| (loss);

\end{tikzpicture}
}
\caption{\textbf{The teacher-student model setup under superposition}, where $K \leq N$. The student learns in a compressed latent space via the embedding layers $\mathbf{W}$ and $\mathbf{W}^\top$. A bias and ReLU nonlinearity are applied to the output to suppress interference noise.}
\label{fig:model_setup}
\end{figure*}
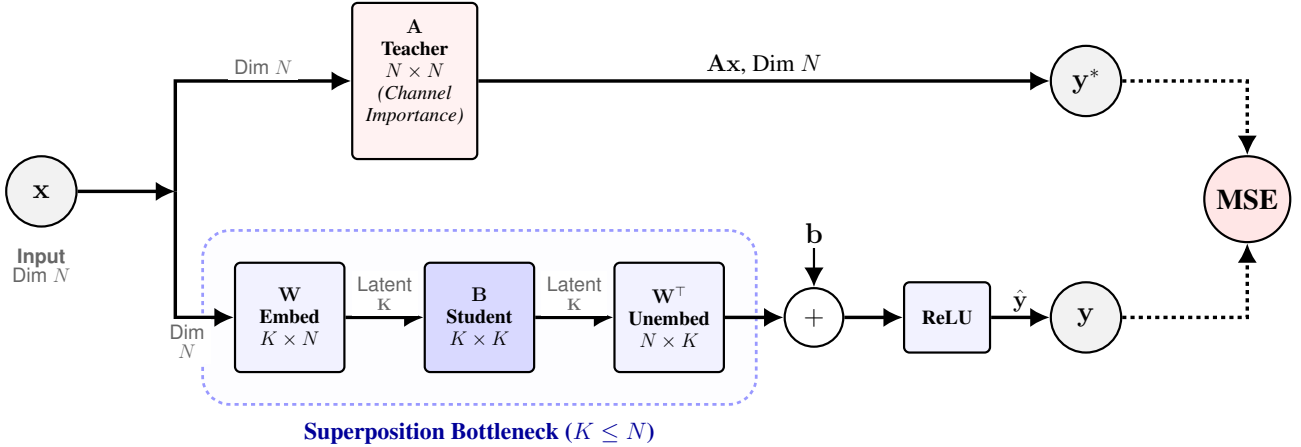

Standard theories of learning often attribute these power laws to the spectral properties of the data \citep{advani2020scaling, bahri2021statistical, sharma2020neural}. In this view, learning is a sequential spectral filtering process where the model fits eigenmodes of the data covariance in descending order of magnitude \citep{saxe2013exact, bordelon2020spectrum, canatar2021spectral}. While this \textit{spectral bias} successfully explains dynamics in linear regimes and kernel methods \citep{simoncini2023spectral, cui2021generalization, bordelon2025feature, bordelon2025icl}, it typically assumes a direct mapping between model dimensions and data features. However, modern LLMs operate in a fundamentally different regime: the latent representations of features are highly \emph{non-orthogonal} of one another, a phenomenon known as \textbf{superposition} \citep{elhage2022toy}.

\textbf{Superposition in LLMs.} 
The dimensionality of the embedding space in LLMs is a critical bottleneck. Models must learn embedding vectors for vocabularies ranging from 32,000 to 256,000 tokens \citep{achiam2023gpt4, team2024gemma} alongside millions of abstract concepts \citep{templeton2024scaling, gao2024scaling}, yet map them into hidden states of only a few thousand dimensions. To surmount this limit, models utilize superposition to store features in non-orthogonal directions \citep{ant2023superposition, bricken2023monosemanticity}. While this increases capacity, it introduces ``interference noise''. Notably, prior works on neural scaling laws often implicitly assume a regime of \textit{sufficient width}, where model dimensions are sufficient to cover the feature space \citep{maloney2022solvable}. This assumption is disconnected from the \textbf{strong superposition} regime of production models \citep{liu2025superposition}, raising the question: how does the interference noise inherent in superposition alter the training dynamics?

To answer this, we must move beyond black-box observation. Just as controlled \textbf{toy models} were essential for isolating the mechanisms of deep linear networks \citep{saxe2019mathematical}, we propose a tractable teacher-student framework to model the training dynamics of LLMs under superposition. Our model strips away architectural complexities of Transformers to focus entirely on the interaction between feature structure and dimensional constraints.

Our investigation yields a surprising divergence in training regimes. In the absence of superposition, the student learns sequentially, with power-law training dynamics strictly determined by input data statistics and channel decay. However, when forced into superposition, the training dynamics becomes unified. The randomness of superposition ``mixes'' features across different channels, effectively equalizing their learning behavior. This leads to a \textbf{universal power-law exponent} of $\alpha \approx 1$, independent of the specific data input and channel decay, while representing an acceleration over the sequential case.

\subsection{Contributions}
Our contributions are as follows:
\begin{itemize}
    \item We introduce a teacher-student toy model that captures the interplay among feature distribution, channel importance, and superposition, reflecting constraints found in production-scale LLMs.
    \item We establish an analytic theory for power-law training dynamics in the \textit{no-superposition} regime, explicitly linking the training exponent to input data statistics and channel importance. 
    \item We demonstrate empirically that the \textit{superposition} regime induces a transition to universal, accelerated power-law training with an exponent $\alpha \approx 1$.
    \item We bridge superposition and scaling laws of training dynamics, proposing that the superposition regime characteristic of LLMs leads to a uniform, accelerated training trajectory compared to the sufficient-width regimes assumed in prior works.
\end{itemize}

\textbf{Organization.} The remainder of this paper is organized as follows. Section \ref{sec:setup} introduces the toy model setup. Section \ref{sec:theory} derives the analytic theory for the no-superposition model. Section \ref{sec:experiment} presents the empirical results for universal acceleration and analyzes the optimal-compute scaling frontiers. Finally, Section \ref{sec:discussion} provides a mechanistic explanation for the universal behavior under superposition. We conclude in Section \ref{sec:conclusion}.

\begin{figure*}[t]
    \centering
    \begin{subfigure}[b]{0.48\textwidth}
        \centering
        \includegraphics[width=\textwidth]{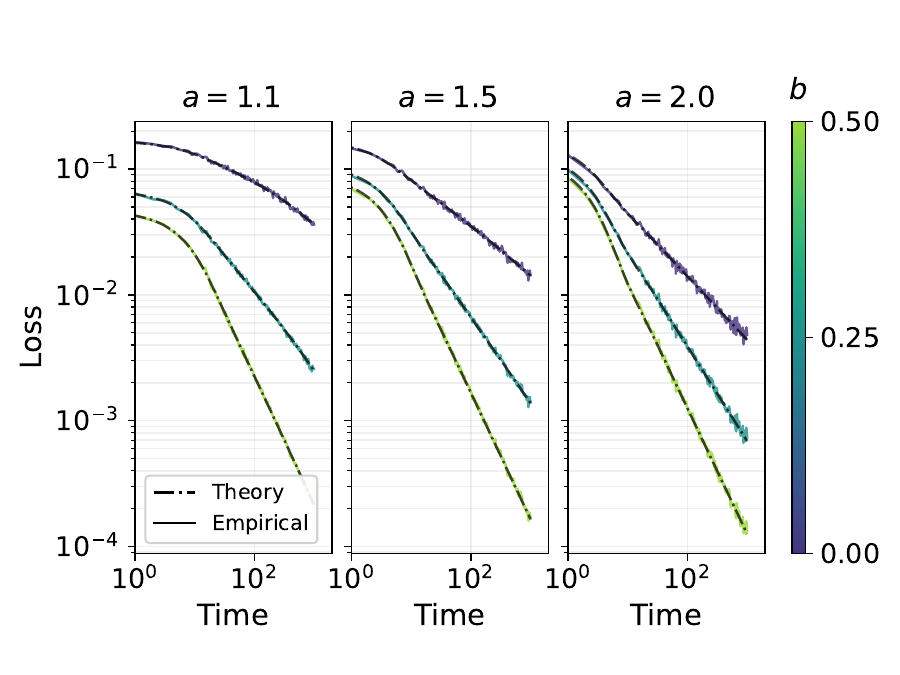}
        \caption{Theory prediction matches the empirical loss trajectory}
        \label{fig:theory_loss}
    \end{subfigure}
    \hfill % Adds horizontal space between subplots
    \begin{subfigure}[b]{0.45\textwidth}
        \centering
        \includegraphics[width=\textwidth]{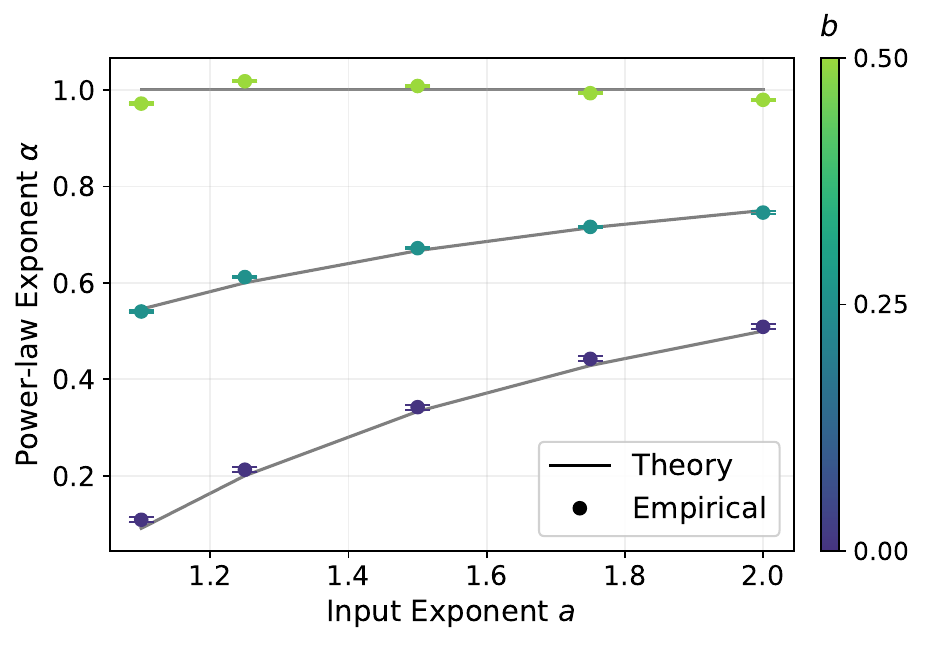}
        \caption{Exponents match $\alpha = (a+2b-1)/a$}
        \label{fig:theory_exponent}
    \end{subfigure}
    \caption{\textbf{Without superposition, learning exponents depend on both input data and channel statistics.} We verify the analytic theory for the no-superposition baseline ($N=K=1024$). \textbf{(a)} Empirical loss curves (solid) track the theoretical predictions (dashed) across varying input decays $a$ and channel importances $b$. \textbf{(b)} The fitted power-law exponents align precisely with the derived scaling law $\alpha = (a+2b-1)/a$, confirming that learning is strictly governed by data and channel statistics.}
    \label{fig:theory_verification}
\end{figure*}

\subsection{Related works}

\textbf{Neural scaling laws.} 
Power-law scaling is well-established for pre-training \citep{kaplan2020scaling, hoffmann2022training}, transfer learning \citep{hernandez2021scaling}, and emergent abilities \citep{wei2022emergent}. Extensions of these laws have explored data pruning \citep{sorscher2022beyond}, mixture-of-experts architectures \citep{clark2022unified}, and state-space models \citep{bi2024mamba}. Theoretical derivations typically rely on kernel methods or infinite-width limits \citep{bordelon2020spectrum, canatar2021spectral}, attributing scaling to the decay of the data manifold's spectrum \citep{sharma2020neural} or the resolution of singularities in the loss landscape \citep{wei2019regularization}. Our work departs from this by investigating the \textit{superposition regime}—where width is the bottleneck—a constraint often absent in kernel-based theories but definitive of LLMs \citep{liu2025superposition}.

\textbf{Theory of learning.} 
A central debate in deep learning theory concerns the distinction between the Neural Tangent Kernel (NTK) regime \citep{jacot2018neural, arora2019exact} and the feature learning regime \citep{chizat2019lazy, yang2021tensor}. While the NTK regime predicts dynamics governed by fixed spectral properties \citep{saxe2013exact, simoncini2023spectral, bordelon2025feature, bordelon2025icl}, LLMs are believed to operate in the feature regime where representations evolve. Our work shows that superposition acts as a bridge: while the student model is linear (like many NTK analyses), the feature mixing induces a collective learning dynamic that breaks the standard spectral linkage found in the NTK regime.

\textbf{Mechanistic interpretability.} 
Mechanistic interpretability aims to reverse-engineer neural networks into understandable components \citep{olah2020zoom}. A key focus is the \textit{superposition hypothesis}, which posits that models represent more features than they have neurons \citep{elhage2022toy}. Recent advances have utilized sparse autoencoders to disentangle these mixed representations \citep{bricken2023monosemanticity, cunningham2023sparse}. However, most studies focus on static, trained networks. Research on the \textit{dynamics} of learning has focused largely on ``grokking'' in algorithmic tasks \citep{power2022grokking, nanda2023progress}. We contribute to this by isolating superposition as a mechanism that actively shapes the continuous scaling laws of training dynamics.

%%% TOY MODEL SETUP %%%

\section{Toy model setup}\label{sec:setup}

To investigate the interplay between feature superposition and training dynamics, we propose a teacher-student framework designed to simulate the feature structure and dimensional constraints characteristic of LLMs.

\subsection{Task definition: data and channels}
We define the learning task via a \textbf{sparse input distribution} and a teacher model, mimicking the multiscale structure of natural language data and Transformer channel importance.

\textbf{Input features.} 
We consider an input vector $\mathbf{x} \in \mathbb{R}^N$ where each component $x_i$ represents a feature activation. To capture the sparsity of natural data, $x_i$ is defined as a mixture of Bernoulli and Uniform distributions:
\begin{equation}
    x_i = u_i v_i, \quad u_i \sim \text{Bernoulli}(p_i), \quad v_i \sim U(0, 1).
\end{equation}
Here, $p_i$ governs feature frequency and follows a power-law decay $p_i \propto i^{-a}$ (with $a > 1$). This distribution ensures that lower-index features are significantly more frequent than higher-index ones. We normalize $p_i$ with an activation density $E = \sum_i p_i = 1$ to maintain consistent activation density across varying $N$.

\textbf{Teacher model.} 
The target signal is generated by a fixed teacher matrix $\mathbf{A} \in \mathbb{R}^{N \times N}$, which models the \textbf{channel importance} observed in Transformer blocks. We impose a power-law spectral decay on the diagonal entries:
\begin{equation}
    A_{ii} = i^{-b}, \quad A_{ij} = 0 \text{ for } i \neq j,
\end{equation}
where $b \geq 0$ is a decay constant. This structure forces the model to prioritize learning lower-index features to minimize loss, establishing an ordered sequence in the optimization landscape.

\subsection{Student architecture and superposition}
The student model $f_{\theta}(\mathbf{x})$ attempts to reconstruct the teacher's output $\mathbf{y}^* = \mathbf{Ax}$ subject to a bottleneck dimension $K$.

\textbf{Superposition and interference.}
As illustrated in Figure \ref{fig:superposition_concept}, when $K < N$, the student must utilize superposition to represent the $N$ features. While this increases capacity, it introduces non-orthogonal interference noise. To model this, the student compresses the sparse input via a fixed, column-normalized random projection (embedding) $\mathbf{W} \in \mathbb{R}^{K \times N}$ into a latent state $\mathbf{h} = \mathbf{W}\mathbf{x} \in \mathbb{R}^K$. The student matrix $\mathbf{B} \in \mathbb{R}^{K \times K}$ processes this latent representation.

The signal is then decoded via the transpose $\mathbf{W}^\top$. Crucially, to manage the interference noise inherent in superposition, we introduce a learnable bias $\mathbf{b}$ and a ReLU nonlinearity at the output:
\begin{equation}
    \mathbf{y} = \text{ReLU}(\mathbf{W}^\top \mathbf{B} \mathbf{W} \mathbf{x} + \mathbf{b}).
\end{equation}
Since the true input signal is non-negative ($x_i \ge 0$), the ReLU function acts as an error-correction mechanism, suppressing negative interference components arising from the non-orthogonal basis.

A graphical illustration of the teacher-student model under superposition is provided in Figure \ref{fig:model_setup}. Note that in the case where \textbf{no superposition} is applied and $K = N$, the embedding layers $\mb{W}$ and $\mb{W}^\top$ are taken as identity matrices. Neither bias nor ReLU nonlinearity is applied in the no-superposition case.

\textbf{Training objective.}
The student is trained to minimize the Mean Squared Error (MSE) relative to the teacher's output. The objective function is:
\begin{equation}
    \mathcal{L} = \frac{1}{2} \mathbb{E}_{\mathbf{x}} \left[ \| \mathbf{y}^* - \mathbf{y} \|_2^2 \right].
\end{equation}
For all empirical experiments, we focus on the same feature dimension $N = 1024$ and omit normalizing the loss by $N$ for simplicity. Under this setup, the dynamics are governed by three parameters: the feature decay $a$, the channel importance decay $b$, and the compression ratio $N/K$.

\begin{figure}[t]
    \centering
    \includegraphics[width=0.45\textwidth]{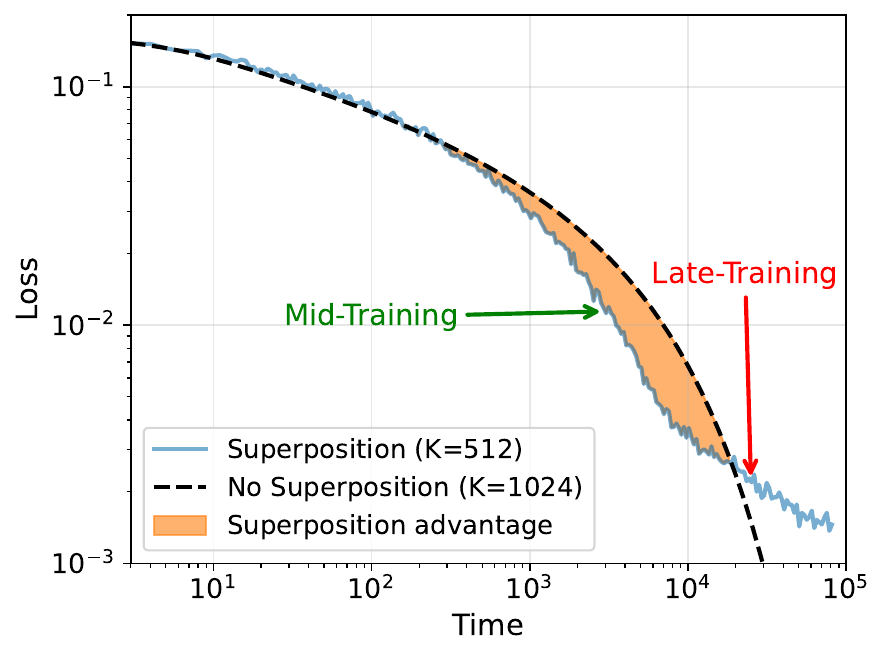}
    \caption{\textbf{Mid-training acceleration via superposition.} Loss from the superposition experiment ($N=1024$, $K = 512$) is compared to the no-superposition theory ($N=K=1024$). A mid-training acceleration in loss convergence appears under superposition despite the bottleneck in $K$.}
    \label{fig:superposition_loss}
\end{figure}

\begin{figure*}[t]
    \centering
    \begin{subfigure}[b]{0.45\textwidth}
        \centering
        \includegraphics[width=\textwidth]{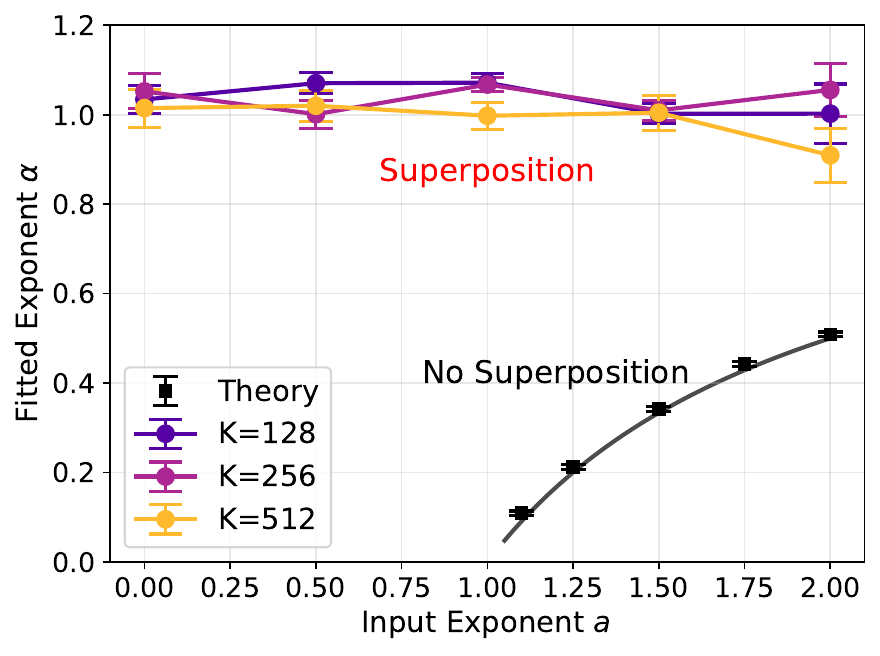}
        \caption{Exponent $\alpha \approx 1$ is independent of input decay $a$}
        \label{fig:a_exponent}
    \end{subfigure}
    \hfill % Adds horizontal space between subplots
    \begin{subfigure}[b]{0.45\textwidth}
        \centering
        \includegraphics[width=\textwidth]{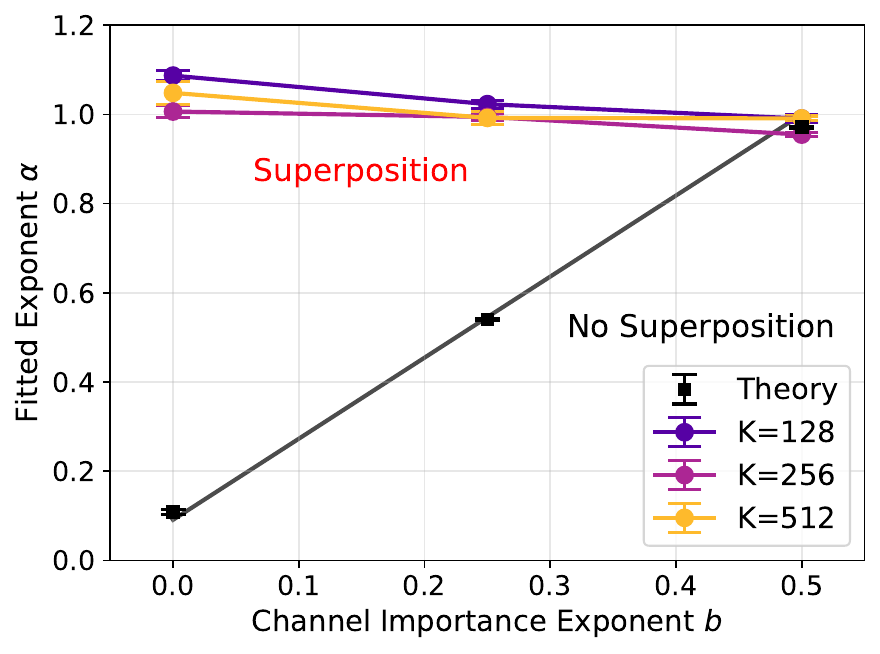}
        \caption{Exponent $\alpha \approx 1$ is independent of channel importance $b$}
        \label{fig:b_exponent}
    \end{subfigure}
    \caption{\textbf{Universality of the training exponent under superposition.} We plot the fitted power-law exponent $\alpha$ for varying student sizes $K \in \{128, 256, 512\}$. Unlike the sequential case where $\alpha$ varies with data and channels, superposition locks the exponent to $\boldsymbol{\alpha \approx 1}$ regardless of the input feature decay $a$ or the channel importance decay $b$.}
    \label{fig:exponent_summary}
\end{figure*}

\begin{figure*}[t]
    \centering
    \includegraphics[width=0.95\textwidth]{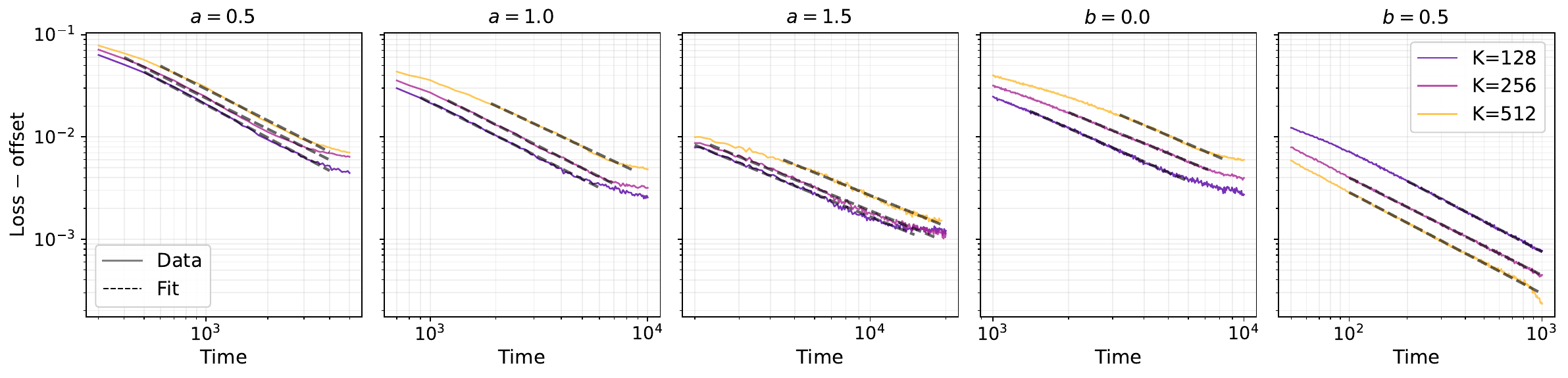}
    \caption{\textbf{Robustness of power-law dynamics.} We plot sample loss curves (solid lines) and their corresponding power-law fits (black dashed lines) for a wide range of input parameters ($a \in [0.5, 1.5], b \in [0, 0.5]$). In all cases, the mid-training dynamics are accurately modeled by a power law with a consistent exponent of $\alpha \approx 1$, confirming the universality of the power-law training exponent.}
    \label{fig:superposition_exp}
\end{figure*}
%%% THEORY %%%

\section{Theory}\label{sec:theory}

In this section, we derive the analytic scaling laws for the training dynamics in the regime without superposition ($N=K, \mathbf{W}=\mathbf{I}$). While the superposition case involves complex interference terms (deferred to Appendix \ref{sec:superposition_theory}), the diagonal case yields a closed-form solution that serves as our baseline. We empirically verify this prediction in Figure \ref{fig:theory_verification}, showing excellent agreement between the theory and experiment.

\subsection{Dynamics without superposition}
In the absence of superposition, the student $\mathbf{B}$ can fully capture the teacher $\mathbf{A}$. Assuming small initialization ($\mathbf{B}(0) \approx 0$), the matrix $\mathbf{B}(t)$ remains diagonal throughout training. The loss decomposes into a sum over independent feature modes:
\begin{equation}
    \mathcal{L}(t) \approx \frac{1}{2} \sum_{i=1}^N \lambda_i (s_i(t) - a_i)^2,
\end{equation}
where $\lambda_i \propto i^{-a}$ are the dominating diagonal variances of the data covariance. $a_i = i^{-b}$ and $s_i(t)$ are the teacher coefficients and student diagonal entries respectively.

In the continuous-time limit with learning rate $\eta$, the dynamics of each diagonal entry $s_i(t)$ follows a linear ODE:
\begin{equation}
    \frac{d s_i}{dt} = \eta \lambda_i (a_i - s_i) \implies s_i(t) = a_i (1 - e^{-\eta \lambda_i t}).
\end{equation}
Substituting this solution back into the loss function yields:
\begin{equation} \label{eq:exact_loss}
    \mathcal{L}(t) = \frac{1}{6} \sum_{i=1}^N i^{-(a+2b)} \exp\left( - \frac{2}{3} \eta t \cdot i^{-a} \right).
\end{equation}
Equation \ref{eq:exact_loss} reveals the spectral filtering nature of gradient descent: features are learned sequentially based on their data frequency $i^{-a}$.
For the rest of the paper, we group the learning rate together with steps and define time as $\eta \cdot t$, and \textbf{denote time simply by $\mb{t}$}.

\subsection{Derivation of the power law exponent}
To extract the asymptotic scaling behavior, we approximate the sum in Equation \ref{eq:exact_loss} with an integral. We define a critical feature index $i_c(t) \propto t^{1/a}$ representing the boundary between learned and unlearned features. Assuming a large feature dimension $N$, the mid-training dynamics (where $1 \ll i_c \ll N$) are dominated by the tail of the unlearned features. As detailed in Appendix \ref{sec:powerlaw_theory}, this integration yields a power-law decay:
\begin{equation} \label{eq:power_law_main}
    \mathcal{L}(t) \propto t^{-\alpha}, \quad \text{where } \alpha = \frac{a + 2b - 1}{a}, \quad (a+2b > 1).
\end{equation}
This result establishes that in the sequential learning regime, the training speed is strictly coupled to the input statistics ($a$) and channel importance ($b$). As $a, b$ increases, the input becomes sparser and channel decays faster, leaving less new information for the student to learn from the teacher, resulting in a faster learning trajectory.

\begin{figure*}[t]
    \centering
    \begin{subfigure}[b]{0.45\textwidth}
        \centering
        \includegraphics[width=\textwidth]{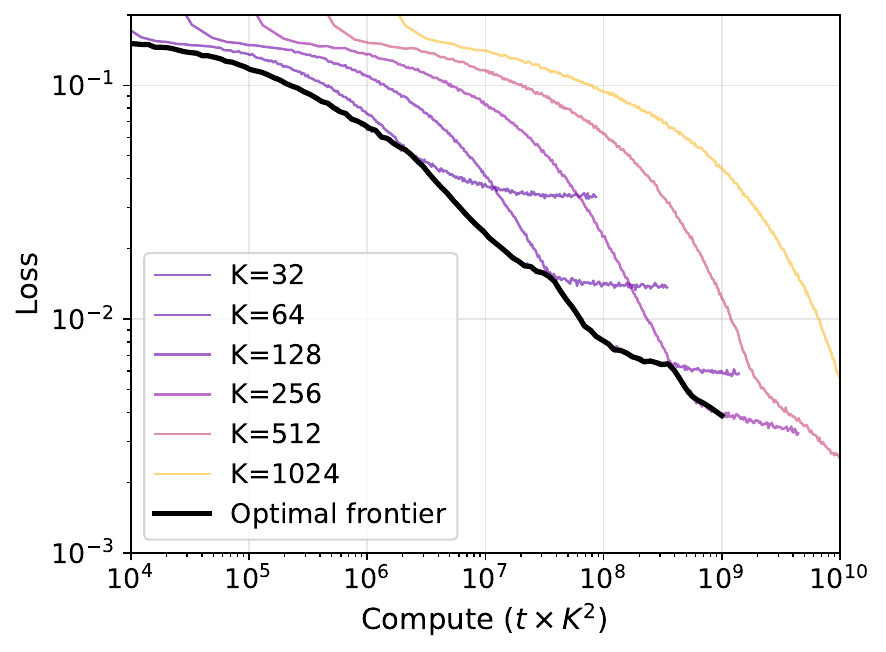}
        \caption{Optimal-compute frontier}
        \label{fig:compute_frontier}
    \end{subfigure}
    \hfill % Adds horizontal space between subplots
    \begin{subfigure}[b]{0.45\textwidth}
        \centering
        \includegraphics[width=\textwidth]{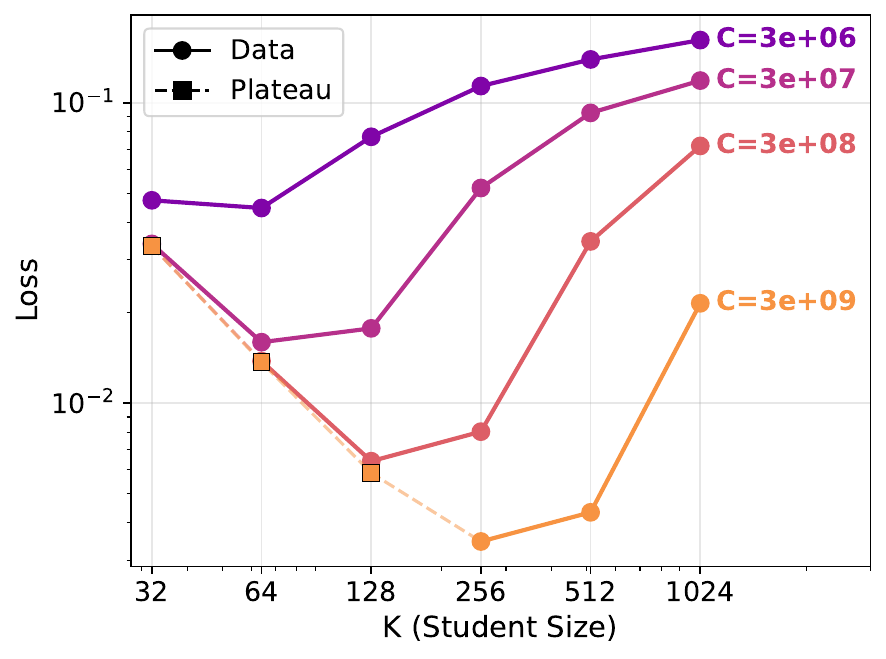}
        \caption{Iso-compute curves identify optimal student sizes}
        \label{fig:compute_isocurves}
    \end{subfigure}
    \caption{\textbf{Demonstration of optimal-compute frontier with model size scaling.} We analyze the trade-off between model size and training duration for a fixed input distribution ($a=1.1, b=0$). \textbf{(a)} The raw training curves (colored lines) are enveloped by a black solid line, defining the optimal-compute frontier. \textbf{(b)} We plot loss against student size $K$ for four fixed compute budgets $C$ (log-uniform). The distinct minima in each curve illustrate the trade-off between $K$ and $C$, mirroring the behavior observed in LLMs.}
    \label{fig:compute_summary}
\end{figure*}

%%% EXPERIMENT %%%

\section{Experiments} \label{sec:experiment}

We now investigate the training dynamics under superposition, where an analytic exponent becomes intractable due to feature mixing. We perform experiments on a toy model of size $N=1024$ trained via SGD with online data generation, mimicking the infinite-data regime of LLM pre-training.

\subsection{Methodology}
To quantify the training dynamics, we adapt the Chinchilla scaling law \citep{hoffmann2022training} to account for the distinct phases of optimization. The standard scaling law models loss as a sum of time-dependent and parameter-dependent terms:
\begin{equation}
    \mathcal{L}(t, K) \approx c_t t^{-\alpha} + c_k K^{-\beta} + \mathcal{L}_0.
\end{equation}

However, in our teacher-student setup, optimization proceeds in two distinct regimes: an early-to-mid phase dominated by optimization error (time-limited), and a late phase dominated by the bottleneck capacity (width-limited), as represented in Figure \ref{fig:superposition_loss}. In our analysis, we focus on the mid-training stage, where $c_t t^{-\alpha} + \mathcal{L}_0$ dominates, to extract the effective training exponent $\alpha$. 

\subsection{Results}
We vary the input decay $a \in [1.1, 2.0]$, channel importance $b \in [0.0, 0.5]$, and student dimension $K \in [32, 1024]$.

\textbf{Theory verification (no superposition).} 
For the baseline case ($K=N$), we confirm our theoretical derivation. As shown in Figure \ref{fig:theory_exponent}, the empirical exponents match the predicted $\alpha = (a+2b-1)/a$ precisely.

\textbf{Superposition regime.}
When the student is forced into superposition ($K < N$), the dynamics undergo a sharp shift. Figure \ref{fig:exponent_summary} summarize the fitted exponents. Note that the requirement of $a + 2b > 1$ is only needed for power-law theory approximation. We can conduct experiments for any $a, b \geq 0$. 

Regardless of the specific values of $a$, $b$, or the compression ratio $K/N$, the mid-training power-law exponent converges to $\boldsymbol{\alpha \approx 1}$. Examples of the loss used for fitting the training exponents are shown in Figure \ref{fig:superposition_exp}.

This represents a universal acceleration compared to the sequential case. For example, with $a=1.1$ and $b=0$, the sequential theory predicts a slow decay of $\alpha \approx 0.09$. Superposition accelerates this by over $\mathbf{10\times}$. This universality suggests that the randomness inherent in the embedding layer acts as a mechanism to equalize the effective learning rates across features, decoupling the optimization dynamics from the spectral decay of the data.

In Appendix \ref{sec:superposition_theory}, we analyze two cases of the superposition model in linear regime: the limits of maximum positive interference and isotropic mixing. They provide a theoretical intuition for the acceleration and uniformity observed in empirical results. However, we stress that they do not account for the $\alpha \approx 1$ exponent, which emerges only with ReLU nonlinearity and bias.

\subsection{Optimal-compute scaling}

While superposition universally accelerates mid-training dynamics regardless of the student size, the final attainable performance of the model is limited by the bottleneck dimension $K$. To understand the trade-offs between model size and training steps, we analyze the optimal-compute frontier.

In Figure \ref{fig:compute_frontier}, we plot the loss curves for students of varying sizes ($K \in [32, 1024]$) against total compute (approximated as $C = t \times K^2$). We observe a clear \textbf{optimal envelope}, indicated by the black line, composed of the minimum loss achievable at any given compute budget. Consistent with observations in large-scale Transformers, the optimal model size increases with the compute budget. As shown in Figure \ref{fig:compute_size}, $K$ doubles roughly as $C$ increases by tenfold. We run a preliminary power-law fitting and obtain a size scaling exponent of $K$ against $C$ at around $0.2702$. 

We further quantify this relationship using iso-compute curves (Figure \ref{fig:compute_isocurves}), which document how loss varies with student size at fixed compute levels. These curves demonstrate that for a fixed compute budget, the minimum loss scales inversely with the student size as a power law. We also discuss experimental results on the loss scaling with student size (i.e. the width-limited loss term $c_k K^{-\beta}$) in Appendix \ref{sec:loss_size}. This confirms that our toy model, despite its simplicity, recapitulates the macroscopic scaling behaviors observed in production LLMs while offering a tractable testbed for understanding the underlying mechanisms.

\begin{figure}[h]
    \centering
    \begin{subfigure}[b]{0.45\textwidth}
        \centering
        \includegraphics[width=\textwidth]{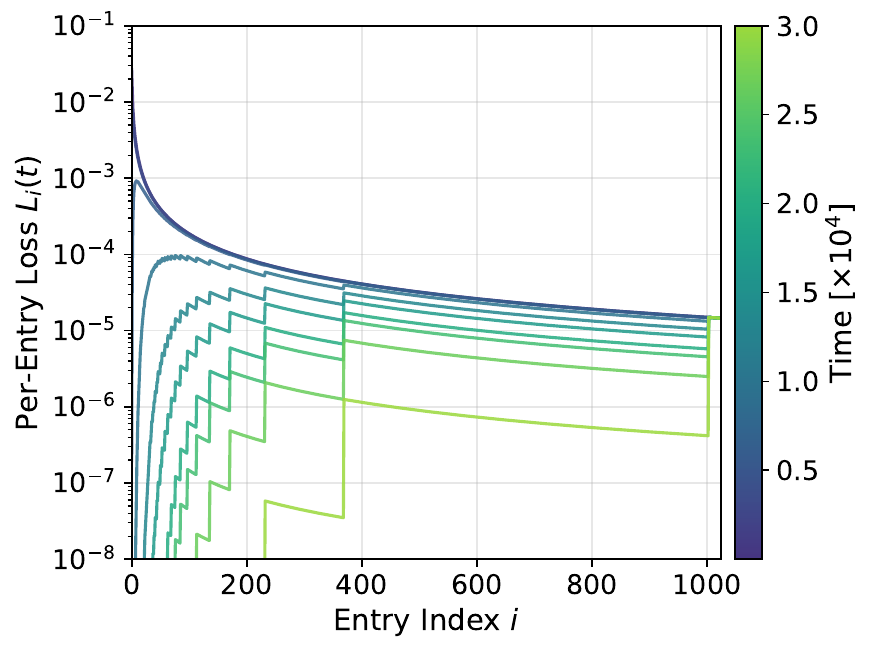}
        \caption{No superposition: sequential ``traveling wave''}
        \label{fig:theory_vector}
    \end{subfigure}
    \hfill
    \begin{subfigure}[b]{0.45\textwidth}
        \centering
        \includegraphics[width=\textwidth]{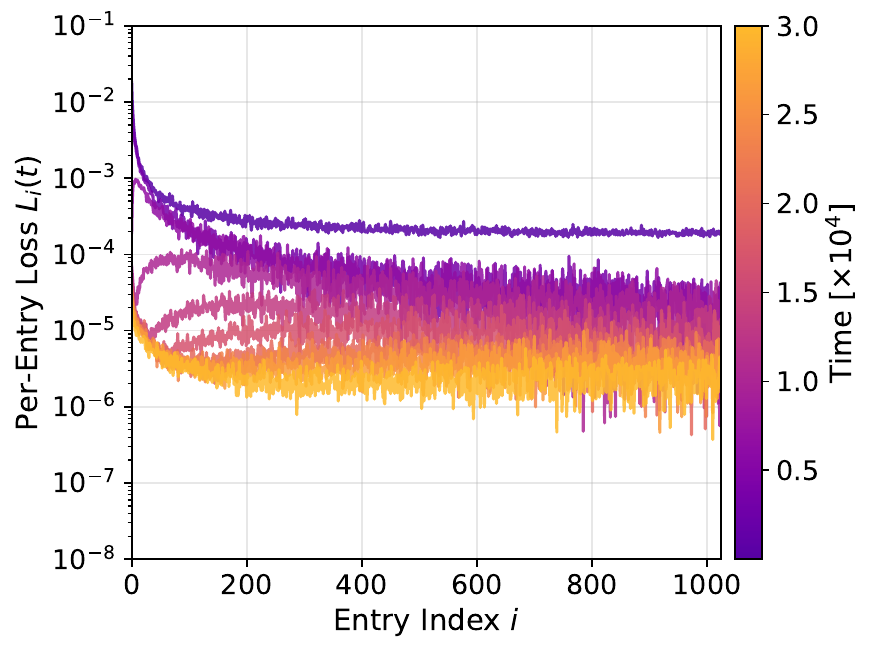}
        \caption{Superposition: parallel ``global decay''}
        \label{fig:superposition_vector}
    \end{subfigure}
    \caption{\textbf{Mechanism of universal loss dynamics under superposition.} We visualize the per-feature loss $L_i(t)$ over time. \textbf{(a)} Without superposition, the model learns features sequentially, creating a traveling wave-front where tail features remain unlearned for long durations. \textbf{(b)} Under superposition, feature mixing equalizes the effective gradients, causing all features—regardless of importance—to be learned in parallel.}
    \label{fig:dynamics_comparison}
\end{figure}

\begin{figure*}[t]
    \centering
    \includegraphics[width=0.7\textwidth]{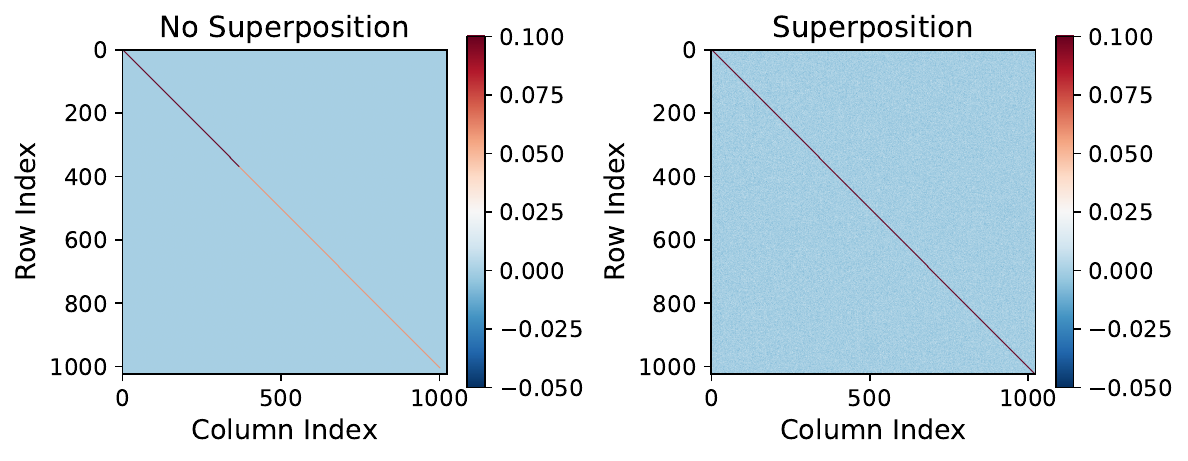}
    \caption{\textbf{Visualizing the student model learning pattern: sequential (no superposition) vs parallel (superposition).} Snapshots of the student matrix $\mathbf{B}$ at $t=10^3$. \textbf{Left:} In the absence of superposition, the student learns strictly along the diagonal, solving features sequentially (only the first $\sim 400$ are learned). \textbf{Right:} The superposition student distributes weights across the entire matrix, learning all 1024 features in parallel.}
    \label{fig:matrix}
\end{figure*}

%%% DISCUSSION %%%

\section{Discussion} \label{sec:discussion}

The central finding of our work is the universal acceleration of training dynamics under superposition. In this section, we investigate the microscopic mechanism driving this phenomenon. Specifically, we address two entangled questions: (1) why is the exponent increased during mid-training, and (2) why is this behavior universal across different data distributions?

The answer lies in how the model processes the input data and channel statistics. We contrast the sequentiality of the baseline case with the parallelism induced by superposition.

\subsection{Mechanism}

\textbf{Sequential learning as a traveling wave.}
In the absence of superposition, the input probability $p_i$ and channel importance $A_{ii}$ follow strict power-law decays by design. This structure forces the gradient descent dynamics to respect an ordered sequence: the model learns features in descending order of importance. 

This phenomenon is illustrated in Figure \ref{fig:theory_vector}, where we plot the loss contribution of individual entries over time in the absence of superposition. The dynamics resemble a ``traveling wave-front''. Features with lower indices (high importance/frequency) are learned first, while features with higher indices (low importance/frequency) remain effectively frozen—their losses do not decrease until the prior, more dominant features are resolved. Consequently, at any point $t$ in mid-training, the total loss is dominated by the accumulated error of the vast number of unlearned features in the tail of the distribution. The rate of loss convergence is thus strictly determined by the spectral decay of the data.

\textbf{Superposition disrupts the sequential order.}
Superposition, via the random embedding $\mathbf{W}$, disrupts this sequence. The compression mixes features from different positions, while normalization per feature ensures they are projected with comparable magnitude. This leads to two fundamental changes:
\begin{enumerate}
    \item \textbf{Feature mixing:} Distinct entries are blended into the latent space, making it impossible for the student to isolate and learn them sequentially.
    \item \textbf{Frequency equalization:} On average, every dimension of the student matrix $\mathbf{B}$ receives a signal that aggregates high-frequency and low-frequency features alike.
\end{enumerate}

As a result, the student learns all entries simultaneously at a comparable rate. This is demonstrated in Figure \ref{fig:superposition_vector}. Unlike the ``traveling wave'' without superposition, the per-entry losses in the superposition regime remain clustered at similar levels and decay in unison. The concept of ``unlearned tail features'' effectively vanishes.

\textbf{Mid-training acceleration.}
Why does this uniformity under superposition lead to acceleration? We emphasize that this advantage is specific to the \textit{mid-training} regime. Since the superposition student ($K < N$) has strictly lower capacity than the teacher ($N$), it must eventually hit a non-zero loss plateau, whereas the ideal student without superposition ($K=N$) converges to zero.

However, before reaching this saturation point, superposition exhibits an advantage. As seen in Figure \ref{fig:superposition_loss}, the superposition loss curve dips significantly lower than the non-superposition one during mid-training. Mathematically, this occurs because the sequential model is penalized by the heavy tail of the power law—it has zero error on learned features but maximal error on the unlearned majority. In contrast, the superposition model distributes the error budget evenly. In the mid-training regime, the sum of many small, averaged errors (superposition) is lower than the sum of the unlearned spectral tail (sequential). The randomness effectively acts to equalize convergence rates, allowing the model to bypass the slow sequential traversal of the spectrum.

\textbf{Visualizing the student structure.}
We provide direct visual confirmation of this structural shift in Figure \ref{fig:matrix}, which displays snapshots of the student matrix $\mathbf{B}$ at time $t=10^3$ (mid-training). Since the teacher is diagonal, an ideal student should recover diagonal entries. 

Without superposition, the student learns strictly along the diagonal, entry-by-entry. At $t=10^3$, it has successfully learned $\sim 400$ of the 1024 features, with the rest remaining at initialization values. With superposition, the student matrix is fully activated. The learned weights are distributed across the entire model to decode the mixed signals. The model effectively learns all 1024 entries in parallel, leveraging the interference in superposition to minimize the global error rate faster than what the sequential structure permits.

%%% CONCLUSION %%%

\section{Conclusion and outlook} \label{sec:conclusion}

In this work, we have established a link between feature superposition and a universal power-law training dynamics. By analyzing a teacher-student framework that mimics the structure of LLMs, we identified two distinct patterns of learning. In the absence of superposition, training is governed by a spectral filtering process, where the loss convergence is determined by the sequential learning of features with decreasing frequencies.

In contrast, we discovered that the introduction of superposition induces a universal change in the dynamics, regardless of input data and channel distributions. The randomness inherent in superposition unifies the effective convergence rates across features, collapsing data-dependent scaling laws into one with a \textbf{universal power-law exponent of $\alpha \approx 1$}.

This finding offers a provocative perspective on the role of model width in training dynamics. Rather than being merely a capacity constraint, a narrow bottleneck—when coupled with high-dimensional feature sparsity—acts to equalize learning dynamics. It trades the precision of orthogonal representation for the speed of parallelized error reduction, providing possible acceleration critical for the efficient training of large-scale models.

\textbf{Outlook.} 
Our findings open several avenues for future research. First, while our toy model captures the linear and feed-forward aspects of superposition, extending this analysis to \textbf{multi-layer architectures with attention mechanisms} is crucial. We hypothesize that the mixing effect of attention heads may further amplify the uniformity we observed. Second, the \textbf{origin of the $1/t$ scaling law} remains mostly elusive. While our linear theory in the Appendix \ref{sec:superposition_theory} provides intuition for the uniformity of convergence rates, a rigorous study of why the exponent settles exactly at $\alpha \approx 1$—rather than other values—remains an open challenge for non-linear superposition models. Third, we focused on the mid-training regime where optimization error dominates. A more granular analysis of the \textbf{late-training dynamics}—where the model hits the irreducible approximation error of the bottleneck—could yield insights into the ``grokking'' phenomena observed in algorithmic tasks. Finally, our results suggest that \textbf{randomness is universally beneficial} for mid-training dynamics. Investigating whether structured or learned embeddings can outperform random projections in the mid or late-training stage remains an open question for optimizing efficient models. We present some initial studies on this subject in Appendix \ref{sec:other_layer}.


% Authors are kindly asked to make their submissions as accessible as possible
% for everyone including people with disabilities and sensory or neurological
% differences. Tips of how to achieve this and what to pay attention to will be
% provided on the conference website \url{http://icml.cc/}.

% If a paper is accepted, we strongly encourage the publication of software and
% data with the camera-ready version of the paper whenever appropriate. This can
% be done by including a URL in the camera-ready copy. However, \textbf{do not}
% include URLs that reveal your institution or identity in your submission for
% review. Instead, provide an anonymous URL or upload the material as
% ``Supplementary Material'' into the OpenReview reviewing system. Note that
% reviewers are not required to look at this material when writing their review.

% \section*{Software and Data}

% We provide relevant codes in a zip file under supplementary materials for anonymous review.

% % Acknowledgements should only appear in the accepted version.
\section*{Acknowledgements}
This work was supported by the Schmidt Polymath Award. ZJC acknowledges support from Kurt Forrest Foundation
Fellowship and Henry Kendall Fellowship. The authors would like to thank the MIT Office of Research Computing and Data for providing access to the Engaging Cluster. We also acknowledge the use of the Della cluster at Princeton University, which is managed by the Princeton Institute for Computational Science and Engineering (PICSciE) and the Office of Information Technology's Research Computing.

The authors acknowledge support from the National Science Foundation under Cooperative Agreement PHY-2019786 (The NSF AI Institute for Artificial Intelligence and Fundamental Interactions). The research was sponsored by the United States Air Force Research Laboratory and the Department of the Air Force Artificial Intelligence Accelerator and was accomplished under Cooperative Agreement Number FA8750-19-2-1000. The computations in this
paper were partly run on the FASRC cluster supported by the FAS Division of Science Research Computing Group at Harvard University. This research used the DeltaAI advanced computing and data resource, which is supported by the National Science Foundation (award OAC 2320345) and the State of Illinois, through allocation CIS240904 from the Advanced Cyberinfrastructure Coordination Ecosystem: Services $\&$ Support (ACCESS) program, supported by National Science Foundation grants $\#$2138259, $\#$2138286, $\#$2138307, $\#$2137603, and $\#$2138296, and through the National Artificial Intelligence Research Resource (NAIRR) Pilot NAIRR250043.

% If a paper is accepted, the final camera-ready version can (and usually should)
% include acknowledgements.  Such acknowledgements should be placed at the end of
% the section, in an unnumbered section that does not count towards the paper
% page limit. Typically, this will include thanks to reviewers who gave useful
% comments, to colleagues who contributed to the ideas, and to funding agencies
% and corporate sponsors that provided financial support.

\section*{Impact Statement}

% Authors are \textbf{required} to include a statement of the potential broader
% impact of their work, including its ethical aspects and future societal
% consequences. This statement should be in an unnumbered section at the end of
% the paper (co-located with Acknowledgements -- the two may appear in either
% order, but both must be before References), and does not count toward the paper
% page limit. In many cases, where the ethical impacts and expected societal
% implications are those that are well established when advancing the field of
% Machine Learning, substantial discussion is not required, and a simple
% statement such as the following will suffice:

% ``This paper presents work whose goal is to advance the field of Machine
% Learning. There are many potential societal consequences of our work, none
% which we feel must be specifically highlighted here.''

% The above statement can be used verbatim in such cases, but we encourage
% authors to think about whether there is content which does warrant further
% discussion, as this statement will be apparent if the paper is later flagged
% for ethics review.

This paper presents a theoretical investigation into the training dynamics of neural networks under superposition. As foundational research, its primary impact is on the scientific understanding of why over-parameterized and compressed models (like LLMs) scale efficiently.

\textbf{Energy and Efficiency.} 
By identifying that superposition induces a universal, accelerated learning trajectory ($t^{-1}$), our work provides a theoretical basis for designing more compute-efficient architectures. Understanding that compressed bottlenecks can accelerate mid-training convergence—rather than hinder it—may encourage the development of narrower, deeper models that maximize this "randomness edge," potentially reducing the carbon footprint and energy costs associated with pre-training foundation models.

\textbf{Interpretability and Reliability.} 
Our work bridges the gap between scaling laws (macroscopic behavior) and mechanistic interpretability (microscopic structure). By elucidating how interference noise actively shapes the training trajectory, we contribute to a better understanding of the internal representations of black-box models. This theoretical grounding is essential for developing more robust interpretability tools, which are strictly necessary for the safe deployment of AI systems in critical domains.

There are no direct negative societal consequences anticipated from this work, though improvements in training efficiency naturally accelerate the general capabilities of AI systems.


\bibliography{refs}
\bibliographystyle{icml2026}

%%%%%%%%%%%%%%%%%%%%%%%%%%%%%%%%%%%%%%%%%%%%%%%%%%%%%%%%%%%%%%%%%%%%%%%%%%%%%%%
%%%%%%%%%%%%%%%%%%%%%%%%%%%%%%%%%%%%%%%%%%%%%%%%%%%%%%%%%%%%%%%%%%%%%%%%%%%%%%%
% APPENDIX
%%%%%%%%%%%%%%%%%%%%%%%%%%%%%%%%%%%%%%%%%%%%%%%%%%%%%%%%%%%%%%%%%%%%%%%%%%%%%%%
%%%%%%%%%%%%%%%%%%%%%%%%%%%%%%%%%%%%%%%%%%%%%%%%%%%%%%%%%%%%%%%%%%%%%%%%%%%%%%%
\newpage
\appendix
\onecolumn

\section{Appendix} \label{sec:appendix}

\subsection{Power-law theory for general distributions} \label{sec:powerlaw_theory}

\begin{figure}[h]
    \centering
    \begin{tikzpicture}[scale=1.2, >=stealth]
        % Axes
        \draw[->, thick] (0,0) -- (6,0) node[right] {Input Decay Parameter $a$};
        \draw[->, thick] (0,0) -- (0,4.0) node[above] {Training Exponent $\alpha$};
        
        % Guidelines
        \draw[dashed, gray] (0,1.5) -- (6,1.5) node[right, gray] {Exponential: $\alpha \approx 1$};
        \draw[dashed, gray] (1,0) -- (1,4.0);
        \node[below] at (1,0) {$a = 1$};
        
        % Curve 1: Power Law Input (a > 1) -> alpha = 1 - 1/a
        % 1.5 corresponds to alpha=1 in this scale.
        \draw[domain=1.2:5.5, smooth, variable=\x, blue, very thick] plot ({\x}, {1.5 * (1 - 1/\x)});
        \node[blue, right] at (3.5, 0.9) {Power Law: $\alpha = 1 - \frac{1}{a}$};
        
        % Curve 2: Algebraic Decay (a > 0) -> alpha = 1 + 1/a
        % "Lower on the left": Start at a=0.6 (y=4) instead of 0.4 (y=5.25)
        % "Extend longer to the right": Extend to a=5.5
        \draw[domain=0.6:5.5, smooth, variable=\x, red, very thick] plot ({\x}, {1.5 * (1 + 1/\x)});
        \node[red, right] at (0.6, 3.8) {Algebraic: $\alpha = 1 + \frac{1}{a}$};
        
        % % Label for Exponential
        % \node[black, align=center] at (5.4, 1.45) {Exponential: };
        
    \end{tikzpicture}
    \caption{\textbf{Phase diagram of training dynamics.} The training power-law exponent $\alpha$ is plotted against the input distribution decay parameter $a$. \textbf{Red:} Inputs with algebraic decay at a finite edge converge fastest ($\alpha > 1$). \textbf{Blue:} Inputs with heavy power-law tails converge slowest ($\alpha < 1$). \textbf{Dashed:} Exponentially decaying inputs (and superposition models) converge at $\alpha \approx 1$.}
    \label{fig:phase_diagram}
\end{figure}
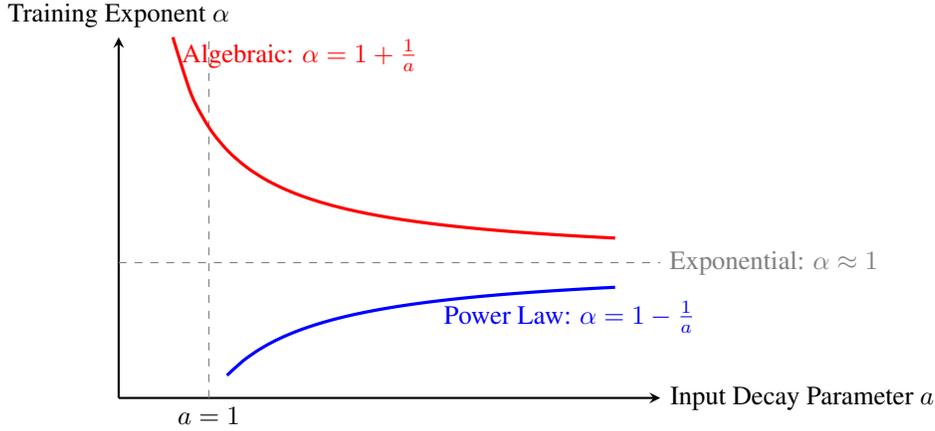

In this section, we provide the detailed derivation of the training dynamics exponents for general input distributions. We broaden the discussion beyond the power-law case analyzed in the main text to include exponential and algebraic decays. We summarize these regimes in the phase diagram in Figure \ref{fig:phase_diagram}.

\subsubsection{Continuous Approximation}
Assuming near-uniform channel importance (in the limit $b \rightarrow 0$), the loss function can be approximated by a continuous integral in the limit of $N \gg 1$. Let $i$ be the discrete feature index and $z$ be its continuous counterpart. The loss is:
\begin{equation}
    \mathcal{L}(t) \propto \sum_{i=1}^N f(i) e^{-f(i) t/ \tau} \approx \int_0^\infty f(z) e^{-f(z) t /\tau} dz,
\end{equation}
where $f(z)$ is the monotone decreasing data frequency. We define a \textbf{critical feature index} $z_c(t)$ where the exponent is of order unity:
\begin{equation}
    f(z_c) \cdot \frac{t}{\tau} = 1 \implies z_c(t) = f^{-1}\left(\frac{\tau}{t}\right).
\end{equation}
At time $t$, features with $z < z_c$ are learned ($e^{-f(z)t/\tau} \approx 0$), while features with $z > z_c$ contribute to the loss ($e^{-f(z)t/\tau} \approx 1$). The loss is dominated by the tail integral:
\begin{equation}
    \mathcal{L}(t) \sim \int_{z_c(t)}^{\infty} f(z) dz.
\end{equation}
Note that for this approximation to be valid, $1 \ll z_c \ll N$, corresponding to the mid-training stage in the dynamics. 

\subsubsection{Power-Law Decay}
Let $f(z) = C z^{-a}$ with $a > 1$. The critical index scales as:
\begin{equation}
    z^{-a}_c \sim \frac{\tau}{t} \implies z_c(t) \sim \left( \frac{t}{\tau} \right)^{1/a}.
\end{equation}
The loss integral becomes:
\begin{equation}
    \mathcal{L}(t) \sim \int_{z_c}^\infty z^{-a} dz = \left[ \frac{z^{1-a}}{1-a} \right]_{z_c}^\infty \propto z_c^{1-a}.
\end{equation}
Substituting $z_c(t)$:
\begin{equation}
    \mathcal{L}(t) \propto \left( t^{1/a} \right)^{1-a} = t^{-(1 - 1/a)}.
\end{equation}

Note that for a nontrivial channel decay with $b > 0$, we can substitute $z^{-a} \rightarrow z^{-(a+2b)}$. The critical feature index $z_c(t)$ only depends on the input data and remains unchanged. Under this substitution, we recover the loss power law shown in Equation \ref{eq:power_law_main}. 

\subsubsection{Exponential Decay}
Let $f(z) = C e^{-\kappa z^a}$ with $\kappa, a > 0$. We perform a change of variables $u = f(z)$. Then $z = f^{-1}(u) = \kappa^{-1/a} (\log(C/u))^{1/a}$.
The differential is:
\begin{equation}
    dz = \frac{d}{du} \left( \kappa^{-1/a} (\log(C/u))^{1/a} \right) du \sim -\frac{1}{a u} (\log(C/u))^{1/a - 1} du.
\end{equation}
Substituting this into the integral $\int f(z) e^{-f(z)t/\tau} dz$:
\begin{equation}
    \mathcal{L}(t) \sim \int_0^C u e^{-(t/\tau)u} \cdot \frac{1}{u} \left( \log \frac{C}{u} \right)^{1/a - 1} du.
\end{equation}
We rescale $s = (t/\tau)u$, so $du = (\tau/t)ds$. In the limit $t \gg 1$, the integral is dominated by small $u$ (finite $s$), so $\log(C/u) = \log(Ct/\tau s) \approx \log(t)$. Pulling the log term out:
\begin{equation}
    \mathcal{L}(t) \sim \left( \log t \right)^{1/a - 1} \int_0^\infty e^{-s} ds \cdot \frac{\tau}{t}.
\end{equation}
Thus, the scaling is:
\begin{equation}
    \mathcal{L}(t) \propto t^{-1} (\log t)^{\frac{1}{a} - 1}.
\end{equation}
The dominant term is $t^{-1}$, representing the decay rate for light-tailed distributions.

\subsubsection{Algebraic Decay}
Let $f(z) \sim (z_* - z)^a$ near a finite edge $z_*$. $z$ approaches $z_*$ from the left: $z \to z^-_*$. The critical index condition $(z_* - z_c)^a \sim \tau/t$ implies the gap $\Delta z = z_* - z_c \sim t^{-1/a}$.
The loss integral is over the gap $\Delta z$:
\begin{equation}
    \mathcal{L}(t) \sim \int_{z_c}^{z_*} (z_* - z)^a dz \propto (z_* - z_c)^{a+1} \sim (t^{-1/a})^{a+1} = t^{-(1 + 1/a)}.
\end{equation}

\subsubsection{Derivation Summary}

\textbf{1. Power-law decay:} 
For $f(z) \sim z^{-a}$ (where $a>1$), the heavy tail of the data slows down learning. 
\begin{equation}
    \mathcal{L}(t) \sim t^{-(1-1/a)}.
\end{equation}
As shown in Fig. \ref{fig:phase_diagram} (blue curve), $\alpha$ approaches 1 only as $a \to \infty$ (steep decay).

\textbf{2. Exponential decay:}
For $f(z) \sim e^{-\kappa z^a}$, the distribution decays fast enough that the dynamics are dominated by the $\mathcal{O}(1)$ time constant.
\begin{equation}
    \mathcal{L}(t) \sim t^{-1} (\log t)^\delta \implies \alpha \approx 1.
\end{equation}
This represents the dividing line in our phase diagram.

\textbf{3. Algebraic decay:}
For distributions with finite support ending at $z_*$, decaying as $(z_* - z)^a$, the scarcity of "hard" examples vanishes.
\begin{equation}
    \mathcal{L}(t) \sim t^{-(1+1/a)}.
\end{equation}
This yields fast convergence ($\alpha > 1$), shown in red in Fig. \ref{fig:phase_diagram}.

\subsubsection{Power-law Verification}

\begin{figure*}[h]
    \centering
    \begin{subfigure}[b]{0.45\textwidth}
        \centering
        \includegraphics[width=\textwidth]{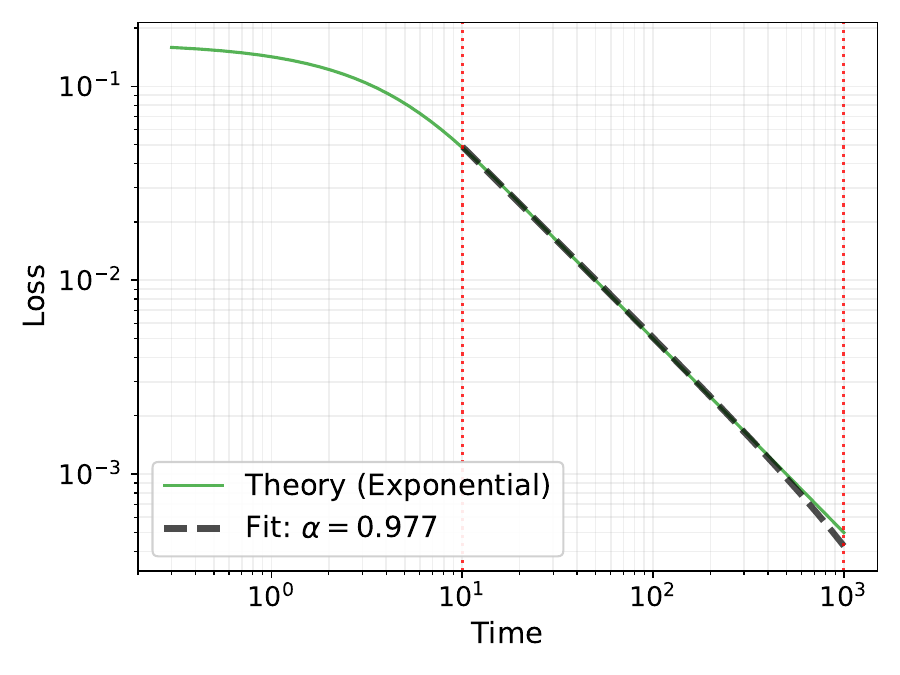}
        \caption{Exponential Decay ($p_i \propto e^{-0.5i}$)}
        \label{fig:exp_fit}
    \end{subfigure}
    \hfill
    \begin{subfigure}[b]{0.45\textwidth}
        \centering
        \includegraphics[width=\textwidth]{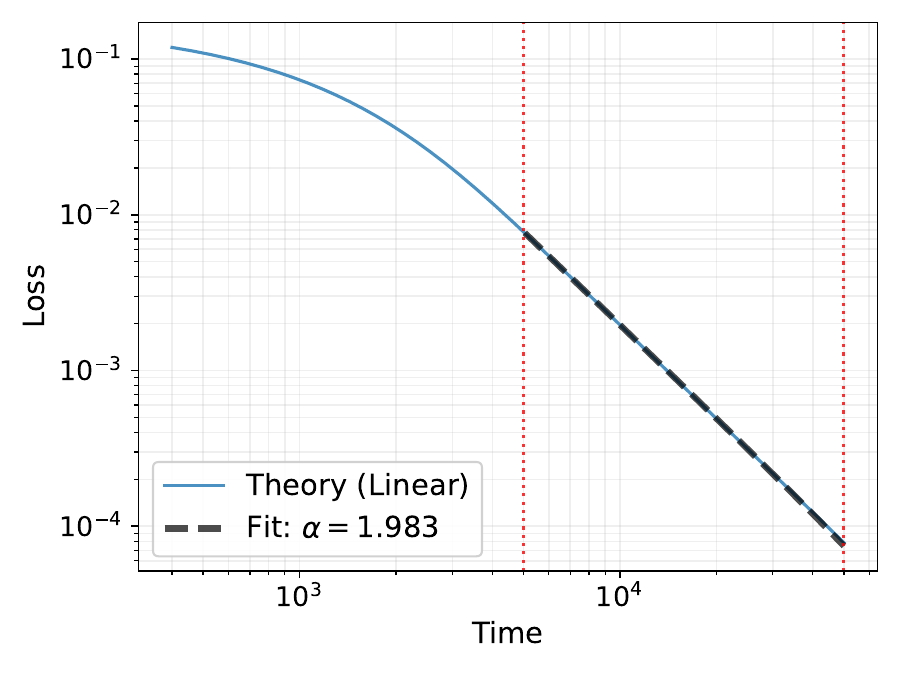}
        \caption{Linear Algebraic Decay ($p_i \propto N-i$)}
        \label{fig:linear_fit}
    \end{subfigure}
    \caption{\textbf{Verification of derived scaling laws for general distributions.} The solid curves show the exact discrete theoretical loss, while the dashed lines show the power-law fit. \textbf{(a)} Exponential input distribution yields $\alpha \approx 1$. \textbf{(b)} Linear algebraic input distribution yields $\alpha \approx 2$. Both match the predictions from our continuous approximation theory.}
    \label{fig:theory_fit_summary}
\end{figure*}

We verify our power-law approximation against the theoretical loss (modifications based on Equation \ref{eq:exact_loss}) for non-power-law distributions. More specifically, we test an exponential decay and a linear algebraic decay. We define the normalized feature frequencies $f(i) = p_i$, normalizing them such that $\sum_i p_i = 1$.  
\begin{itemize}
    \item Exponential decay: $p_i \propto e^{-0.5 i}$. Here, $a = 1$ in the exponential-decay class, and the power-law approximation predicts $\mathcal{L}(t) \propto t^{-1}$, i.e. $\alpha = 1$. 
    \item Linear algebraic decay: $p_i \propto N - i$. Here, $a = 1$ in the algebraic-decay class, and the power-law approximation predicts $\mathcal{L} \propto t^{-(1+1/1)} = t^{-2}$, i.e. $\alpha = 2$. 
\end{itemize}

In Figure \ref{fig:theory_fit_summary}, we fit power-law training exponents to the exact discrete-sum theoretical loss. For the exponential case (Figure \ref{fig:exp_fit}), we obtain a fitted exponent of $\alpha = 0.9769 \pm 0.0012$, closely matching the approximate prediction of $\alpha = 1$. For the linear decay case (Figure \ref{fig:linear_fit}), we obtain $\alpha = 1.9834 \pm 0.0007$, in excellent agreement with the theoretical prediction of $\alpha = 2$. These results confirm that our continuous integral approximation (Section \ref{sec:powerlaw_theory}) accurately captures the discrete training dynamics across distinct universality classes.

\subsection{Theoretical loss for the linear superposition model} \label{sec:superposition_theory}

We derive the analytical loss dynamics for the superposition model in the linear regime. We analyze how the structure of the embedding matrix $\mathbf{W}$ influences the effective learning dynamics by considering two limiting cases: (1) $\mathbf{W}$ is a designed cluster maximizing positive interference, and (2) $\mathbf{W}$ is composed of random vectors in the isotropic limit of $N \gg K$. Specifically, limit (1) elucidates the mechanism of \textit{acceleration}, while limit (2) explains the emergence of \textit{universality}. The empirical superposition experiments presented in the main text lie between these two extremes, inheriting both the speed of constructive interference and the uniformity of isotropic mixing.

\subsubsection{General Setup}
We consider a linear student $\mathbf{y} = \mathbf{W}^\top \mathbf{B} \mathbf{W} \mathbf{x}$ trying to mimic a linear teacher $\mathbf{y}^* = \mathbf{A} \mathbf{x}$. The loss is:
\begin{equation}
    \mathcal{L} = \frac{1}{2} \text{Tr} \left[ (\mathbf{W}^\top \mathbf{B} \mathbf{W} - \mathbf{A})^\top (\mathbf{W}^\top \mathbf{B} \mathbf{W} - \mathbf{A}) \mathbf{\Sigma} \right],
\end{equation}
where $\mathbf{\Sigma} = \mathbb{E}[\mathbf{x}\mathbf{x}^\top] \approx \text{diag}(\sigma_1^2, \dots, \sigma_N^2)$ is the data covariance, dominated by the diagonal variances. Defining the projection covariance $\mathbf{C} \equiv \mathbf{W} \mathbf{\Sigma} \mathbf{W}^\top$ and the correlation matrix $\mathbf{P} \equiv \mathbf{W} \mathbf{W}^\top$, the loss simplifies to:
\begin{equation}
    \mathcal{L} = \frac{1}{2} \left[ \text{Tr}(\mathbf{A} \mathbf{\Sigma} \mathbf{A}) + \text{Tr}(\mathbf{B}^\top \mathbf{P} \mathbf{B} \mathbf{C}) - 2 \text{Tr}(\mathbf{S} \mathbf{B}) \right],
\end{equation}
where $\mathbf{S} = \mathbf{W} \mathbf{\Sigma} \mathbf{A} \mathbf{W}^\top$ represents the projected teacher signal.
The gradient flow dynamics for the student matrix $\mathbf{B}$ are given by:
\begin{equation} \label{eq:matrix_ode}
    \frac{d\mathbf{B}}{dt} = -\eta \frac{\partial \mathcal{L}}{\partial \mathbf{B}} = -\eta (\mathbf{P} \mathbf{B} \mathbf{C} + \mathbf{C} \mathbf{B} \mathbf{P} - \mathbf{S} - \mathbf{S}^\top) / 2.
\end{equation}
For symmetric matrices, we analyze the leading order dynamics: $\dot{\mathbf{B}} \approx -\eta (\mathbf{P} \mathbf{B} \mathbf{C} - \mathbf{S})$.

\subsubsection{Case 1: Clustered Superposition}
To analyze the acceleration mechanism, we construct a structured embedding. We partition the $N$ features into $K$ clusters $\{G_1, \dots, G_K\}$. We assume an idealized "hard" assignment where each feature $i \in G_j$ is mapped perfectly to dimension $j$: 
\begin{equation}
    W_{ji} = \begin{cases} 1 & \text{if } i \in G_j \\ 0 & \text{otherwise} \end{cases}.
\end{equation}
In this limit, the matrices $\mathbf{P}$ and $\mathbf{C}$ become diagonal:
\begin{equation}
    \mathbf{P} = \text{diag}(n_1, \dots, n_K), \quad \mathbf{C} = \text{diag}\left(\sum_{i \in G_1} \sigma_i^2, \dots, \sum_{i \in G_K} \sigma_i^2\right),
\end{equation}
where $n_j = |G_j|$ is the cluster size. The teacher projection $\mathbf{S}$ is also diagonal with entries $S_{jj} = \sum_{i \in G_j} A_{ii} \sigma_i^2$.
In this symmetric, diagonal limit, the ODE decouples for the diagonal entries $b_j$ of $\mathbf{B}$:
\begin{equation}
    \dot{b}_j \approx -\eta (n_j C_{jj} b_j - S_{jj}).
\end{equation}
This linear ODE has the solution:
\begin{equation}
    b_j(t) = \frac{S_{jj}}{n_j C_{jj}} \left( 1 - e^{-\eta (n_j C_{jj}) t} \right).
\end{equation}
The solution describes an exponential relaxation. Crucially, the effective convergence rate for cluster $j$ is $\lambda_{eff} \propto n_j C_{jj} = n_j \sum_{i \in G_j} \sigma_i^2$. This explicitly shows that clustering accelerates learning by summing the variances (importance) of all features in the cluster.

\textbf{Interpretation:} This scenario represents an extreme limit of superposition characterized by maximum positive interference. This is analogous to the "privileged basis" discussed in \citet{elhage2022toy}. By perfectly aligning features into the same subspace, the model achieves two multiplicative acceleration effects: (i) the summation of signal energy ($\sum \sigma^2$) and (ii) the accumulation of gradients from $n_j$ features. This results in a quadratic acceleration scaling $\sim n_j^2 \bar{\sigma}^2$, representing a theoretical upper bound of training speed, albeit at the cost of high collision error.

\subsubsection{Case 2: Isotropic Limit}
We now consider the isotropic limit where $\mathbf{W} \in \mathbb{R}^{K \times N}$ is a random matrix with $N \gg K$. We assume the embedding is column-normalized such that the $N$ feature vectors are unit length $\|\mathbf{w}_i\|^2 = 1$. This corresponds to entries $W_{ji} \sim \mathcal{N}(0, 1/K)$.
Due to the isotropy of random vectors, the correlation matrix $\mathbf{P}$ becomes a scaled identity matrix determined by the frame potential:
\begin{equation}
    \mathbb{E}[\mathbf{W} \mathbf{W}^\top] = \sum_{i=1}^N \mathbb{E}[\mathbf{w}_i \mathbf{w}_i^\top] = \frac{N}{K} \mathbf{I}_K.
\end{equation}
Next, we consider the projection covariance $\mathbf{C} = \mathbf{W} \mathbf{\Sigma} \mathbf{W}^\top$. With $\mathbf{\Sigma} = \text{diag}(\sigma_i^2)$, the matrix averages the input variance scaled by the compression ratio:
\begin{equation}
    \mathbb{E}[\mathbf{C}] = \sum_{i=1}^N \sigma_i^2 \mathbb{E}[\mathbf{w}_i \mathbf{w}_i^\top] = \frac{1}{K} \left( \sum_{i=1}^N \sigma_i^2 \right) \mathbf{I} = \frac{N}{K} \bar{\sigma}^2 \mathbf{I}.
\end{equation}
Substituting these into the dynamics (Eq. \ref{eq:matrix_ode}):
\begin{equation}
    \frac{d\mathbf{B}}{dt} \approx -\eta \left[ \left(\frac{N}{K}\mathbf{I}\right) \mathbf{B} \left(\frac{N}{K} \bar{\sigma}^2 \mathbf{I}\right) - \mathbf{S} \right] = -\eta \left(\frac{N}{K}\right)^2 \bar{\sigma}^2 \mathbf{B} + \eta \mathbf{S}.
\end{equation}

\textbf{Interpretation:} This represents the maximum uniform/isotropic limit of superposition. This result explains the universality. In the non-superposition case, each feature $i$ learns at its own rate $\lambda_i \propto i^{-a}$, causing a bottleneck at the tail. With random superposition, the randomness acts as a "uniform clustering" of all features. It statistically achieves the same quadratic scaling $(N/K)^2$ as the uniformly clustered case. This spectral unification ensures that the student matrix $\mathbf{B}$ converges uniformly at a single effective rate, eliminating the spectral bottleneck and leading to a universal exponent.

\subsection{Is randomness near-optimal for mid-training?}\label{sec:other_layer}

In our main analysis, we utilized fixed random projections for the embedding layers $\mb{W}$ and $\mb{W}^\top$. A natural question arises: is the universal acceleration observed in mid-training an artifact of fixed randomness, or does it persist when the embeddings are optimized?

To address this, we modify the toy model setup such that both the encoder $\mb{W}$ and decoder $\mb{W}^\top$ are fully learnable parameters. This increases the parameter count from $K^2$ (student only) to $K^2 + 2NK$. We repeat the superposition experiments with input decay $a = 1.1$, channel importance $b = 0.0$, and feature dimension $N = 1024$, sweeping student sizes $K \in \{128, 256, 512\}$. We fit the power-law training exponent in the same mid-training window ($t \sim 10^3$ to $10^4$) used in the fixed-embedding experiments.

\begin{figure*}[h]
    \centering
    \begin{subfigure}[b]{0.45\textwidth}
        \centering
        \includegraphics[width=\textwidth]{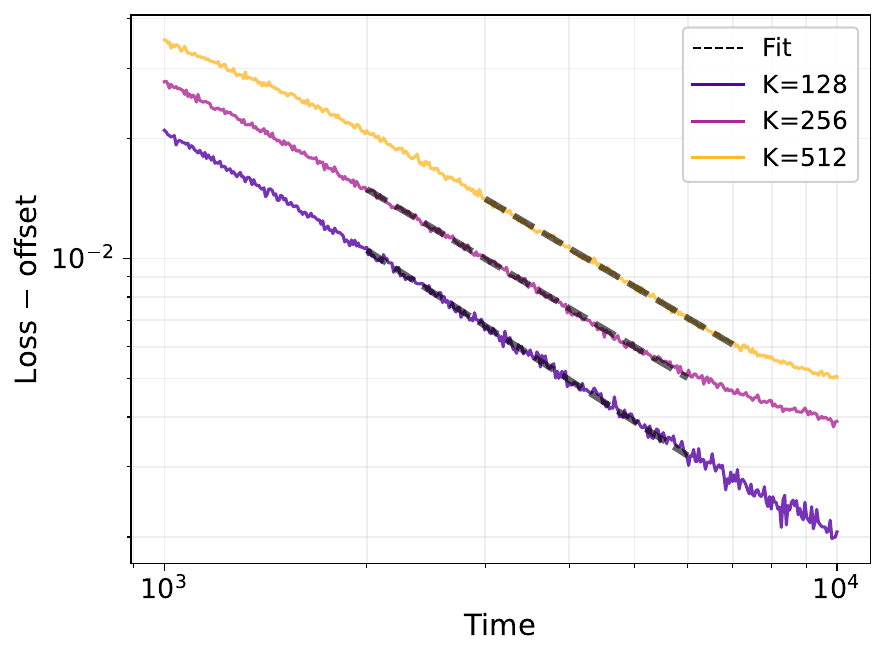}
        \caption{Loss Dynamics (Learnable $\mb{W}$)}
        \label{fig:learnable_loss}
    \end{subfigure}
    \hfill
    \begin{subfigure}[b]{0.45\textwidth}
        \centering
        \includegraphics[width=\textwidth]{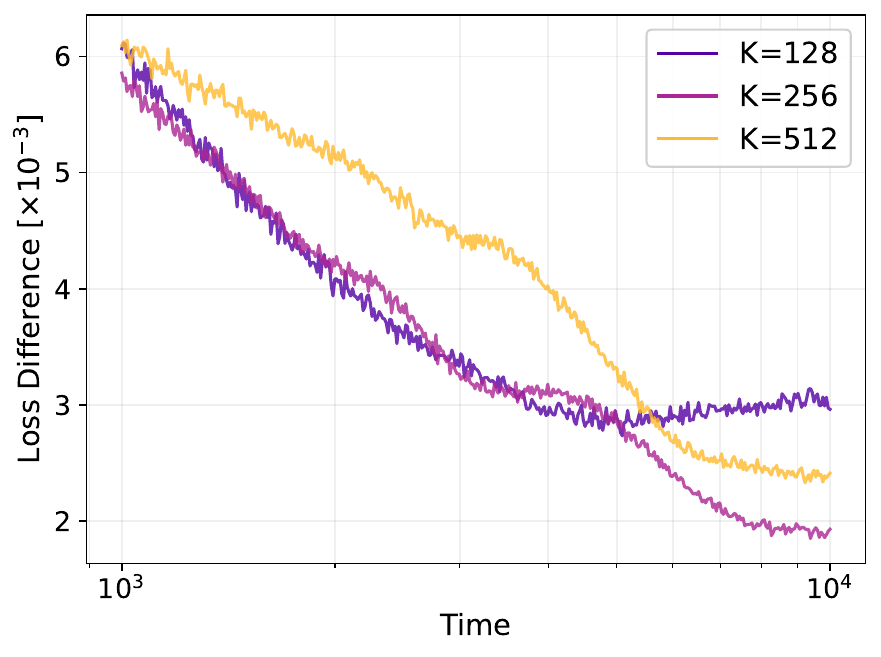}
        \caption{Loss Difference ($\mathcal{L}_{\text{fixed}} - \mathcal{L}_{\text{learnable}}$)}
        \label{fig:learned_difference}
    \end{subfigure}
    \caption{\textbf{Comparison of fixed vs. learnable embeddings.} \textbf{(a)} Training curves for the learnable $\mb{W}$ model show the same $\alpha \approx 1$ power-law decay as the fixed random model. \textbf{(b)} The loss difference between fixed and learnable models. While learnable embeddings achieve slightly lower loss (positive difference), the improvement is an order of magnitude smaller than the total loss, indicating that randomness is the primary driver of the training exponent.}
    \label{fig:learned_summary}
\end{figure*}

\textbf{Results.} 
As shown in Figure \ref{fig:learnable_loss}, we find that the power-law training dynamics remain robust to this architectural change. The fitted exponents for the learnable case are:
\[
\alpha_{128} = 1.0812 \pm 0.0215, \quad \alpha_{256} = 0.9933 \pm 0.0149, \quad \alpha_{512} = 0.9946 \pm 0.0206.
\]
These values are indistinguishable from the $\alpha \approx 1$ regime observed with fixed random embeddings. This suggests that for mid-training dynamics, the "mixing" provided by random initialization is sufficient to induce the universal acceleration effect; explicit gradient optimization of the basis directions does not further accelerate the \textit{rate} of loss convergence (i.e. the power =-law training exponent).

\textbf{Performance gap.} 
While the exponent remains unchanged, learnable embeddings do offer a performance advantage. In Figure \ref{fig:learned_difference}, we plot the loss difference $\Delta \mathcal{L} = \mathcal{L}_{\text{fixed}} - \mathcal{L}_{\text{learnable}}$. The positive difference indicates that the learnable model achieves a lower absolute loss. However, the magnitude of this improvement is $\mathcal{O}(10^{-3})$, which is a secondary effect compared to the absolute loss magnitude of $\mathcal{O}(10^{-2})$. 

We conclude that randomness is near-optimal for the mid-training scaling laws. We emphasize, however, that this claim applies specifically to the mid-training regime; given infinite training time, the learnable model will naturally converge to a lower final loss plateau due to its increased capacity and ability to align the bottleneck with the principal components of the data.

\subsection{Width-limited loss scaling}\label{sec:loss_size}

For completeness, we analyze the scaling of the width-limited loss term $\mathcal{L}_{t \to \infty}(K) \sim K^{-\beta} + \mathcal{L_0}$, which dictates the irreducible error floor at the end of training due to model capacity.

We investigate the relationship between the final saturation loss and the student dimension $K$ using the superposition setup with fixed input and channel statistics ($a = 1.1, b = 0.0$). We sweep the bottleneck size $K$ from $32$ to $512$. To isolate the capacity-limited term, we calculate the average loss from the final training plateau (late-time dynamics), distinct from the mid-training power-law regime analyzed in the main text.

\begin{figure}[h]
    \centering
    \includegraphics[width=0.45\textwidth]{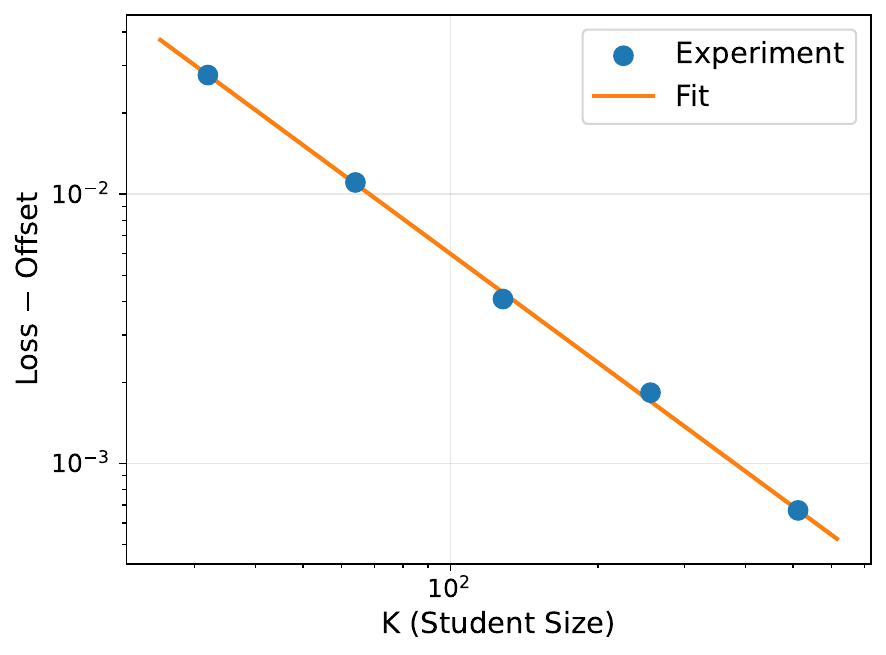}
    \caption{\textbf{Scaling of the final plateau loss against student width $K$.} The loss decays as a power law $K^{-\beta}$ with $\beta \approx 1.3$, consistent with capacity scaling laws in superposition regimes.}
    \label{fig:loss_k}
\end{figure}

As shown in Figure \ref{fig:loss_k}, the final loss follows a clear power-law scaling with respect to width. A fit to the data yields an exponent of $\beta = 1.30 \pm 0.01$. This result aligns closely with width-scaling laws reported in recent studies on superposition \citep{liu2025superposition}, confirming that in addition to a universal \textit{rate} of learning ($\alpha$), the \textit{capacity} loss of the model scales roughly inversely with the bottleneck dimension $K$.

\begin{figure}[h]
    \centering
    \includegraphics[width=0.45\textwidth]{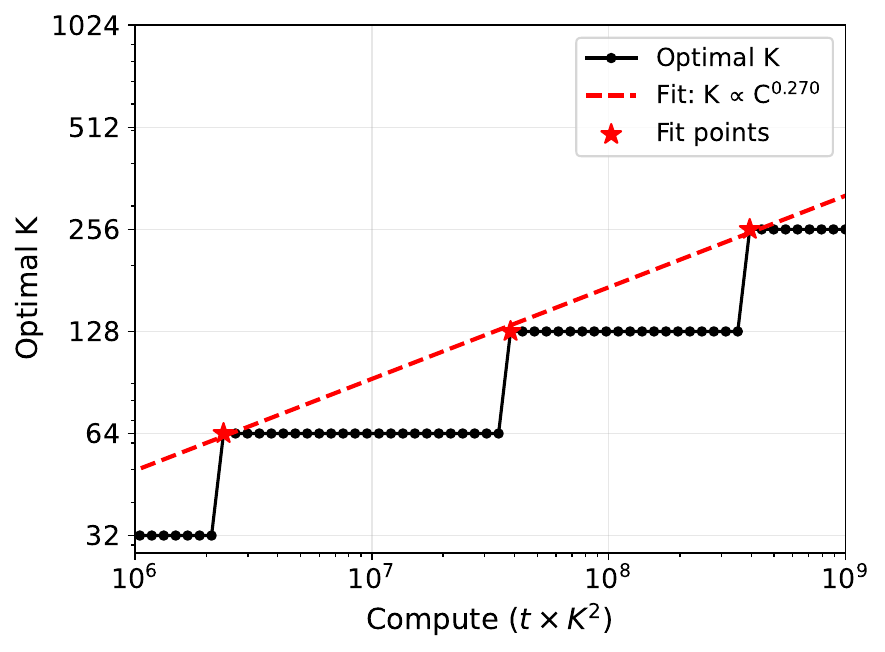}
    \caption{\textbf{Optimal model size scales with compute.} The optimal student size $K$ extracted from the frontier scales with the compute budget, enabling the prediction of optimal resource allocation. Under a rough estimation, the size scaling exponent against compute is $\approx 0.2702$.}
    \label{fig:compute_size}
\end{figure}

 % % \begin{tabular}{ |p{3cm}||p{3cm}|p{3cm}|p{3cm}|p{3cm}|p{3cm}|}
% %  \hline
% %  \multicolumn{5}{|c|}{\textcolor{red}{Title?}} \\
% %  \hline
% %   & $a$ & $K$ & $\alpha$ & $L_0$\\
% %  \hline
% % 0 & 0.0 & 128 & 1.034097 $\pm$0.030416 & 0.005835 \\
% % 1 & 0.0 & 256 & 1.052220  $\pm$0.039086 & -0.001802 \\
% % 2 & 0.0 & 512 & 1.014845 $\pm$0.042492 & -0.007782 \\
% % 3 & 0.5 & 128 & 1.070476 $\pm$0.023256 & 0.003577 \\
% % 4 & 0.5 & 256 & 1.001129 $\pm$0.031103 & -0.001973 \\
% % 5 & 0.5 & 512 & 1.020075 $\pm$0.034901 & -0.003945 \\
% % 6 & 1.0 & 128 & 1.071596 $\pm$0.019746 & 0.003899 \\
% % 7 & 1.0 & 256 & 1.067442 $\pm$0.015866 & 0.000808 \\
% % 8 & 1.0 & 512 & 0.997706 $\pm$0.029830 & -0.001496 \\
% % 9 & 1.5 & 128 & 1.002191 $\pm$0.022340 & 0.001116 \\
% % 10 & 1.5 & 256 & 1.009691 $\pm$0.022029 & 0.000814 \\
% % 11 & 1.5 & 512 & 1.004378 $\pm$0.039295 & 0.000404 \\
% % 12 & 2.0 & 128 & 1.002374 $\pm$0.066242 & 0.000546 \\
% % 13 & 2.0 & 256 & 1.055514 $\pm$0.060000 & 0.000511 \\
% % 14 & 2.0 & 512 & 0.909275 $\pm$0.060000 & 0.000368 \\
% %  \hline
% % \end{tabular}

%%%%%%%%%%%%%%%%%%%%%%%%%%%%%%%%%%%%%%%%%%%%%%%%%%%%%%%%%%%%%%%%%%%%%%%%%%%%%%%
%%%%%%%%%%%%%%%%%%%%%%%%%%%%%%%%%%%%%%%%%%%%%%%%%%%%%%%%%%%%%%%%%%%%%%%%%%%%%%%

\end{document}